\documentclass[10pt,journal,compsoc]{joser1}
\pdfoutput=1

\ifCLASSINFOpdf
   \usepackage[pdftex]{graphicx}
\else
   \usepackage[dvips]{graphicx}
\fi

\usepackage{amssymb}
\usepackage[hang]{subfigure}
\usepackage{microtype}

\journalnumber{1}                       
\journalvolume{5}                       
\journalmonth{May}                
\journalyear{2014}                      
\articlefirstpage{63}                  
\articlelastpage{79}                   
\copyrightauthor{C. Berger}
\headoddname{C. Berger/From a Competition for Self-Driving Miniature Cars to a Standardized Experimental Platform: Concept, Models, Architecture and Evaluation}%

\begin{document}

\title{From a Competition for Self-Driving Miniature Cars to a Standardized Experimental Platform: \\ Concept, Models, Architecture, and Evaluation}

\author{
Christian Berger
\thanks{{\bf Regular paper} -- Manuscript received October 15, 2013;
revised April 30, 2014.}

\IEEEcompsocitemizethanks{\IEEEcompsocthanksitem This work was
supported by Chalmers $|$ University of Gothenburg, Sweden and HiQ AB.\protect\\

\IEEEcompsocthanksitem Authors retain copyright to their papers
and grant JOSER unlimited rights to publish the paper
electronically and in hard copy. Use
of the article is permitted as long as the author(s) and the journal are properly
acknowledged.}

} 

\address{
Department of Computer Science and Engineering, Chalmers $|$ University of Gothenburg, Sweden\\
christian.berger@gu.se
}

\markboth

\IEEEcompsoctitleabstractindextext{%
\begin{abstract}
\textit{Context:} Competitions for self-driving cars facilitated the development
and research in the domain of autonomous vehicles towards potential solutions for
the future mobility.
\textit{Objective:} Miniature vehicles can bridge the gap between
simulation-based evaluations of algorithms relying on simplified models, and
those time-consuming vehicle tests on real-scale proving grounds.
\textit{Method:} This article combines findings from a systematic literature
review, an in-depth analysis of results and technical concepts from contestants
in a competition for self-driving miniature cars, and experiences of
participating in the 2013 competition for self-driving cars.
\textit{Results:} A simulation-based development platform for real-scale
vehicles has been adapted to support the development of a self-driving miniature
car. Furthermore, a standardized platform was designed and realized
to enable research and experiments in the context of future mobility solutions.
\textit{Conclusion:} A clear separation between algorithm conceptualization and
validation in a model-based simulation environment enabled efficient and riskless
experiments and validation. The design of a reusable, low-cost, and energy-efficient
hardware architecture utilizing a standardized software/hardware interface enables
experiments, which would otherwise require resources like a large real-scale test track.
\end{abstract}

}

\maketitle

\section{Introduction}
\label{sec:Introduction}
\IEEEPARstart{S}{elf-driving} vehicle technology as fostered by international
competitions like the DARPA Urban Challenge \cite{RBL+08,BBBC+08} or the
Grand Cooperative Driving Challenge in 2011 is reported to be available for
customers by the end of this decade \cite{Hir13}. On the one hand, this
technology shall replace driving tasks where the human driver is
``under-challenged'', for example long distance travels on highways. Thus,
the driver can focus on other tasks during such periods like doing business or
relaxing. On the other hand, such technology is said to be infallible
from human failure because computer programs never get tired. Thus, they
can manage complex and critical traffic situations where the human driver
might be ``over-challenged'' -- for example unexpectedly crossing 
pedestrians between vehicles parked on a sideways parking strip where
emergency braking is required \cite{LB12}. In such situations, a pattern-based algorithm,
which is classifying the criticality of the current situation perceived from the
surroundings, does not suffer from the ``reaction time'' in contrast to
the human driver. Therefore, vehicles that intelligently monitor the
surroundings are supposed to reduce severely or fatally injured traffic
participants.

\subsection{Motivation}

The aforementioned scenarios assume that the technology, which is used to
realize comfort assistant and safety systems for critical traffic
situations, is fault-tolerant and robust itself. Thus, the driver -- especially when he/she is
``out of the driving loop'' and hence, not mentally involved in the traffic
situation -- relies and depends on the technology, which needs to handle all
traffic situations that could happen regularly during a vehicle's life-time in
a safe and reliable manner. Therefore, the system of a self-driving and
potentially interconnected vehicle needs to be as good and robust as the
ordinarily experienced driver.

However, the question arises how to determine and model the ``average
experience'' in such a way that a self-driving system can make sense out of it.
Furthermore, this experience model depends at least on different countries,
the capability to reliably predict the traffic participants' future behavior, and
in some cases even on the ``eye contact'' -- a level of communication that is
hardly realizable for self-driving cars.

Thus, further research and experiments are required to develop algorithms,
evaluation methodologies, and protocols for information interchange
to understand dependencies between traffic maneuvers and to increase the
future reliability of these systems \cite{Ber12a}. However, the more
complex the traffic scenarios become like autonomous overtaking maneuvers
on highways or turning assistants at intersections, the more space and further
actors are needed to conduct experiments and validations in a repeatable manner
on proving grounds.

Therefore, simulations are used to investigate interactively such traffic
scenarios or to run extensive simulations in an automated manner to ensure
the quality of a software system implementation \cite{BR12a}. Nevertheless,
these simulations rely on models of reality, which in turn base on
simplifications and assumptions. However, transferring results from simulations
to real scale vehicles on a proving ground might bear the risk that technical
challenges like syntactical or semantic incompatibilities on the software/software
or software/hardware level delay the actual experiments. Furthermore, real scale
experiments require enough space to run a test and means to ensure
repeatability. Here, an intermediate step between purely digital simulations and
real scale experiments can help to transfer results from the simulation and to
evaluate real world effects in a safe and yet manageable manner \cite{BDG+13}.

\subsection{Research Questions}

In consequence, this article aims to investigate the following research questions
mainly on the example of the international competition ``CaroloCup'' in
Germany\footnote{www.carolocup.de} for self-driving miniature vehicles:

\begin{description}
\item[\textit{RQ-1:}] \textit{Which design decisions for self-driving miniature vehicles with respect to sensors and algorithms to fulfill the tasks (a) lane-following, (b) overtaking obstacles, (c) intersection handling, and (d) sideways parking have to be considered?}

\item[\textit{RQ-2:}] \textit{Which design decisions in the software and hardware architecture towards a reusable and standardized experimental platform need to be regarded?}

\item[\textit{RQ-3:}] \textit{How can a development and evaluation environment of a real scale self-driving vehicle be adapted and reused during the design, development, and evaluation of a self-driving miniature vehicle?}

\item[\textit{RQ-4:}] \textit{Which design considerations for a reusable and extendable algorithm for autonomous lane-following with overtaking obstacles are of interest?}

\item[\textit{RQ-5:}] \textit{Which design considerations for a reusable and extendable algorithm for autonomous parking on a sideways parking strip need to be considered?}

\end{description}

\subsection{Contributions of the Article}

The contributions of this article are (a) results from a systematic literature
review for relevant work, (b) an analysis of the results of the last
five years from participants in a competition for self-driving miniature
vehicles; (c) the systematic investigation of design decisions from nine
participating teams in the 2013 competition with respect to sensors
and algorithms for supporting the perception and feature handling in the
aforementioned use cases; and (d) derived considerations and design drivers
towards a standardized experimental platform with respect to the
hardware/software interface as well as concepts and evaluations for the
fundamental algorithms for lane-following with overtaking obstacles and
parking on a sideways parking strip.

\subsection{Structure of the Article}

In Sec.~\ref{sec:RelatedWork}, a structured approach to find similar work
with respect to the design criteria for a standardized automotive-like
experimental platform was carried out. Furthermore, a systematic investigation
of the results from the last years' attendees in the competition as well as an
analysis of their respective approaches is conducted. Sec.~\ref{sec:Design}
uses the results from analyzing this state-of-the-art to discuss and
outline considerations about the design for the software and hardware
architecture for a self-driving miniature vehicle experimental platform; a
specific focus is given to experimenting and evaluating algorithms in a
virtual test environment that was already successfully applied to the
development for a real scale self-driving vehicle. In
Sec.~\ref{sec:LaneFollowing} and \ref{sec:Parking}, two reusable and
extendable algorithms to realize the fundamental behavior of a self-driving
vehicle are described and evaluated. Sec.~\ref{sec:BestPractices} summarizes
important design criteria for the software and hardware architecture
of an experimental platform and gives an overview of the software development
and evaluation process, which was applied during the development of the
self-driving miniature car ``Meili'' that participated successfully in the
``Junior Edition'' of the 2013 CaroloCup competition.
In Sec.~\ref{sec:Conclusion}, the article concludes and gives an outlook about
future work.

\section{Related Work}
\label{sec:RelatedWork}

This section investigates the current state-of-the-art related to this work.
Additionally, the results from the international competition for self-driving
miniature cars between 2009 and 2013 are described alongside with an
analysis of the concepts of the competitors from the 2013 edition.

\subsection{Systematic Literature Review}

A systematic literature review (SLR) according to the guidelines of
Kitchenham and Charters \cite{KC07} to address \textit{RQ-2} was conducted.

\subsubsection{Data Sources and Search Strategy}

For the SLR, the following digital libraries were used as data sources:
ACM Digital Library (ACM DL), IEEE Xplore Digital Library (IEEE Xplore),
ScienceDirect, and SpringerLink. As an additional resource, Google Scholar
was used as a meta-search to complement the results of the digital
libraries. Within these databases, the following search strategy was
applied:

\textit{Publication date:} ``2004 or newer'' because the related work of
interest should be influenced by the DARPA Grand Challenges and
beyond.

\textit{Search phrase:} ``design'' AND (``driver'' OR ``decision'') AND
``software architecture'' AND ``hardware architecture'' AND (``reusable''
OR ``reusability'') AND (``miniature'' OR ``small-scale'') AND (``vehicle''
OR ``vehicular'' OR ``mobile'') AND ``platform''

The search phrase could be directly applied at IEEE Xplore, ScienceDirect,
and Google Scholar. For ACM DL and SpringerLink, the search phrase was
unrolled to 24 individual searches, whose results were merged afterwards.

\subsubsection{Study Selection}

After retrieving results from the databases, the following
\textit{exclusion criteria} were applied: (a) The publication was outside the
specified time frame ($E_1$), (b) the publication was a duplicate ($E_2$),
(c) the publication is an ISO standard ($E_3$), or (d) the title and abstract
are referring to a clearly unrelated domain ($E_4$). Afterwards, the
remaining papers were classified as (e) re-evaluation $\circlearrowright$
or (f) directly related ($\surd$). Finally, all results were merged and
duplicates among all databases were removed. The results are shown in
Tab.~\ref{tab:SLR-Results} reflecting that six publications are directly
relevant to this work.

\begin{table}[t!]
\begin{center}
\begin{tabular}{ l || c | c | c | c | c | c | c}
Library & \# & $E_1$ & $E_2$ & $E_3$ & $E_4$ & $\circlearrowright$ & $\surd$ \\
\hline
ACM DL & 46 & 7 & 31 & - & 8 & - & -\\
Google Scholar & 62 & 11 & - & - & 40 & 7 & 4 \\ 
IEEE Xplore & 17 & 3 & - & 6 & 7 & - & 1 \\
ScienceDirect & 5 & 1 & - & - & 3 & 1 & -\\
SpringerLink & 401 & 102 & 203 & - & 78 & 15 & 3 \\
\hline
Total & 531 & 124 & 234 & 6 & 136 & 23 & 8 \\
\hline
\hline
Total (unique) & 156 &  &  &  &  & 22 & 6 \\ 
\end{tabular}
\end{center}
\caption{Results from searching for related work during the different stages of the SLR.\label{tab:SLR-Results}}
\end{table}

\vspace{-\baselineskip}

\subsubsection{Study Quality Assessment}

Before the relevant data was extracted from the identified papers, publications
from group ``$\circlearrowright$'' were re-evaluated. Four papers turned out to
be thematically clearly unrelated to the work, one was a thorough description of
experiences from applying control theory in practice (cf.~\cite{SJPJ13}), and two
were dealing with sensor networks (cf.~\cite{NGH13,PKL09}). Another five papers
from that group were evaluating robotic software frameworks and agent systems
or describing research roadmaps thereof (cf.~\cite{MKMO11,BDR+12,MKMO11a,WPM+05,Wey10}).
When summarizing these findings after removing duplicates, 132 papers were
thematically not relevant based on the applied search phrase and 16 papers are
related to the work outlined in this article.

\subsubsection{Data Extraction and Synthesis}

These remaining 16 papers were further analyzed with respect to (a) clearly
describing \textit{design drivers/decisions} for their presented work (Design),
(b) description of a proposed \textit{software architecture} (SW-A), (c)
description of a proposed \textit{hardware architecture} (HW-A), (d) addressing
\textit{reusability} (Reuse), (e) incorporating \textit{simulative} aspects (Sim), and
targeting the domain of \textit{miniature or small-scale vehicles} (MSV). These
aspects are shown in Tab.~\ref{tab:SLR-Aspect}, where the rows 1-6 show the
results from papers of group ``$\surd$'', the rows 7-13 list relevant
publications of group ``$\circlearrowright$'' after re-evaluation, and the last
three rows show papers targeting \textit{unmanned aerial vehicles} (UAS).

\begin{table}[t!]
\begin{center}
\begin{tabular}{ c || c | c | c | c | c | c }
Ref. & Design & SW-A & HW-A & Reuse & Sim & MSV \\
\hline
\cite{TRW09} & X & X & - & X & X & - \\
\cite{RKB07} & X & - & X & X & - & - \\
\cite{Bae08} & X & X & X & X & X & - \\
\cite{GBV09} & X & X & X & X & - & X \\
\cite{MJSC08} & X & X & X & X & - & - \\
\cite{SEG+09} & X & - & - & - & X & - \\
\hline
\cite{Wre08} & X & X & - & X & X & - \\
\cite{KGM06} & X & X & - & - & - & - \\
\cite{CB11} & X & X & - & X & - & - \\
\cite{Smu07} & X & X & - & X & X & - \\
\cite{Kon06} & X & X & X & X & - & - \\
\cite{Col10} & X & X & - & - & - & - \\
\cite{Gar10} & X & X & - & - & - & \\
\hline
\cite{GS08} & X & X & X & - & X & - \\
\cite{HWV03} & X & X & - & X & X & - \\
\cite{SL08} & X & - & X & X & - & -\\
\end{tabular}
\end{center}
\caption{Classifying important aspects from related work.\label{tab:SLR-Aspect}}
\end{table}

The synthesized results in Tab.~\ref{tab:SLR-Aspect} show that the addressed topics
of developing and maintaining an appropriate architecture enabling reusability
is an important design consideration for the software and hardware. A clear
incorporation of a simulative approach in the development process is only
reported by approx.~38\% from the first and second group of papers; however, it
clearly plays an important role during the development of algorithms and systems
for UAS.

\subsection{Systematic Competition Analysis}

To address \textit{RQ-1}, both the results from the last five years of the
international competition for miniature self-driving cars described in
Sec.~\ref{sec:CaroloCup}--\ref{sec:CaroloCupParking} and individual technical
concepts of the 2013 contestants were investigated. Therefore, the
concept presentations of the teams published after the competition on
the competition's website were analyzed and classified as reported in the
concept classification matrix in Sec.~\ref{sec:TechnicalConceptMapping}.

\begin{figure}[t!]
  \begin{center}
	\includegraphics[width=.9\linewidth,trim=0cm 2cm 1cm 2cm,clip]{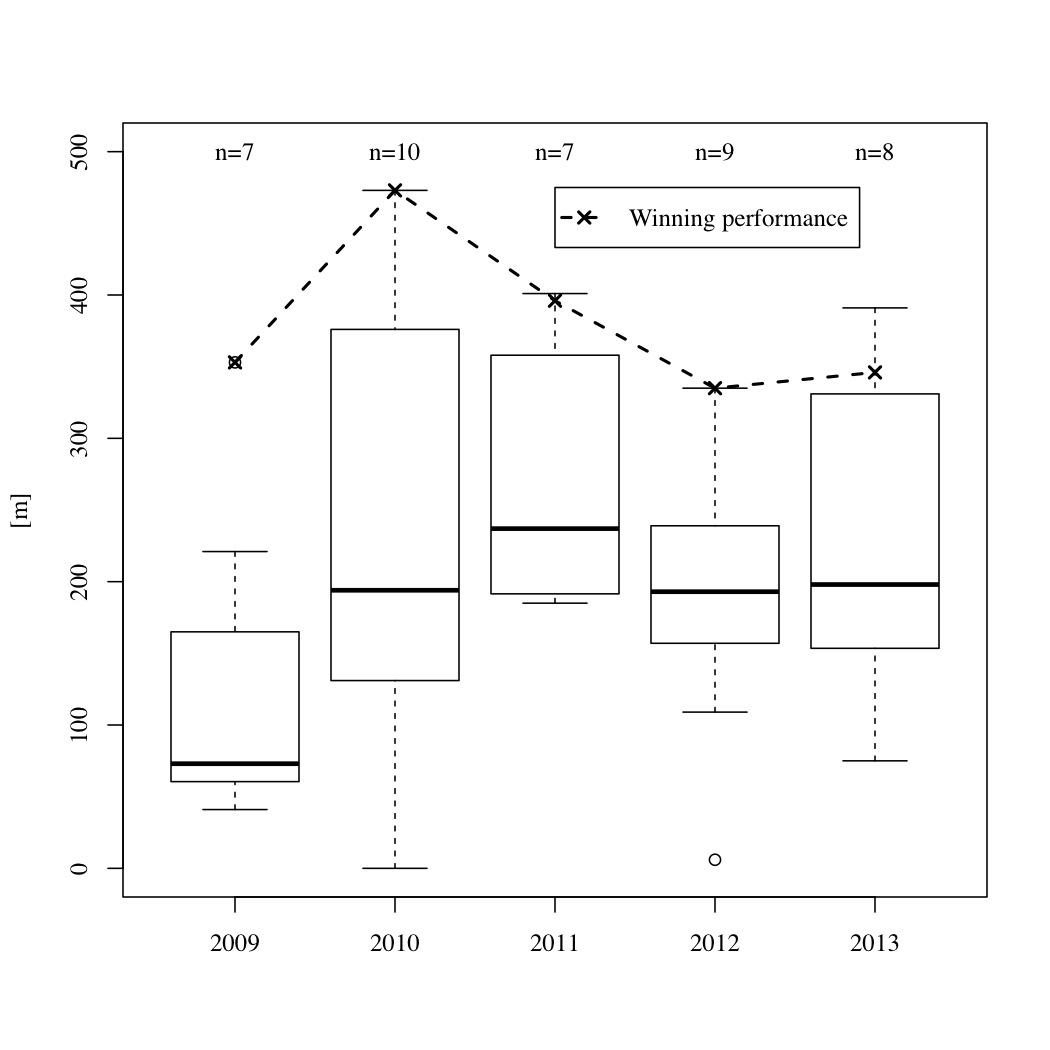}
        \caption{Distribution of driven distances on the track without obstacles and without penalty meters of the last five competitions.}
        \label{fig:Task1}
  \end{center}
\end{figure}

\subsubsection{Competition ``CaroloCup''}
\label{sec:CaroloCup}

The international competition ``CaroloCup'' is an annual competition for
self-driving miniature vehicles founded in 2008 after the DARPA Urban
Challenge and carried out at TU Braunschweig in Germany. The goal is to
facilitate education and research related to self-driving vehicular
technology with a focus on urban environments. Participants have to develop
a 1/10 scale autonomously driving vehicle consisting of hardware and software
that is able to (a) follow a track made out of lane markings, (b) behave correctly
at intersections and overtake stationary and dynamic obstacles, and (c) park
autonomously on a sideways parking strip according to the official rules and
regulations document \cite{CC12}.

Using this approach with a competitive character, participants get in contact
with algorithmic challenges of this technology, which is supposed to be part
of the future intelligent cars as described in Sec.~\ref{sec:Introduction}.
Especially the aspects of developing, testing, integrating, and mastering a
complex distributed and embedded system that has to reliably process volatile
data perceived from the surroundings by various sensors is similar to today's
approaches for self-driving cars \cite{BR12b}.

The aforementioned use cases to be covered by contestants need to be realized
with a certain level of quality in terms of adherence to these requirements to avoid

\begin{enumerate}
\item Penalties when using the remote control system due to vehicle malfunction.
\item Penalties when leaving the regular track.
\item Penalties when colliding with obstacles.
\item Penalties when not positioning the vehicle correctly in the parking gap.
\item Penalties when not obeying traffic rules like yielding right-of-way or misuse of direction and braking lights.
\end{enumerate}

\subsubsection{Task: ``Lane-Following''}

The first task for participants is to drive as many rounds as possible within
3min on a previously unknown track consisting of black ground and white
lane markings. Fig.~\ref{fig:Task1} shows the distribution of the driven distances
on the track without obstacles and without penalty meters according
to the aforementioned list. Obviously, the distance of driven meters on
average has increased from 2009 to 2011 significantly. In the last two
editions, it was around 200m while the interquartile range was the
smallest in 2012.

\begin{figure}[t!]
  \begin{center}
        \includegraphics[width=.9\linewidth,trim=0cm 2cm 1cm 2cm,clip]{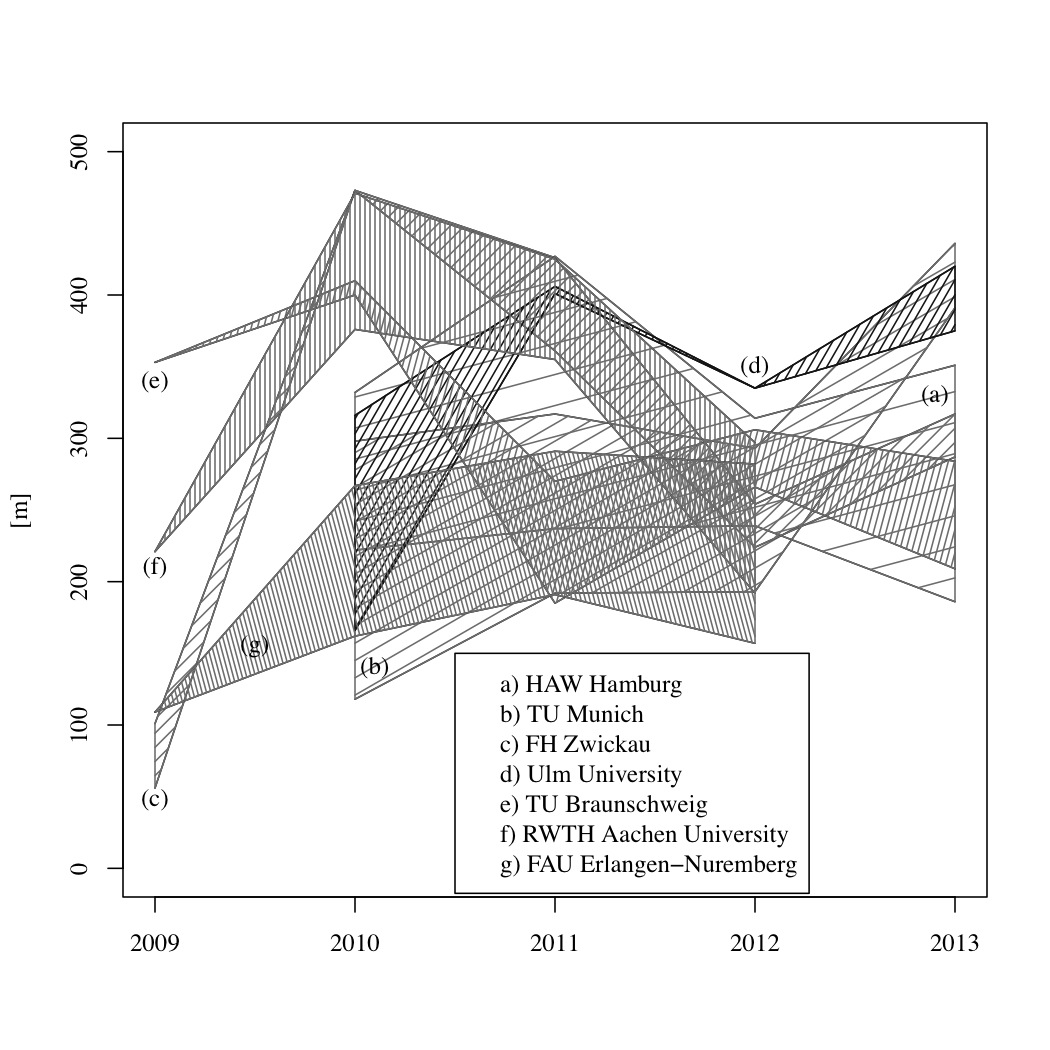}
        \caption{Visualization of driven distances including penalty meters of participants, who participated at least four times in the last five years.}
        \label{fig:Task1LongTerm}
  \end{center}  
\end{figure}

However, Fig.~\ref{fig:Task1} does not take the penalty meter
distribution over the last years into account. Thus, Fig.~\ref{fig:Task1LongTerm}
shows the individually driven distances including the received penalty meters
from those teams who participated at least four times in the last five years.
The polygons per team over time show on the top edge the totally driven
distance and on the bottom edge the distance reduced by penalty meters.
Thus, the less both edges are apart from each other, the more reliable is the
developed system from a team. This can be seen for the team from FH Zwickau,
Ulm University, and TU Braunschweig for example. Furthermore, it is obvious
that the team from Ulm University constantly improved the quality of their
vehicle because the amount of penalty meters is shrinking from 2010 until 2012.
Additionally, they improved not only the overall quality in this driving task but
they also increased the amount of the driven distance to win this
discipline in 2012. The median velocity for these teams was $1.59\frac{m}{s}$ and
for the top team $2.17\frac{m}{s}$ in 2013. Furthermore, this chart also
indicates that several contestants accepted a more risky driving behavior in
the 2010 competition, which resulted in a much higher best performance
compared to the other competitions.

\subsubsection{Task: ``Lane-Following with Obstacle Overtaking''}

\begin{figure}[t!]
  \begin{center}
        \includegraphics[width=\linewidth,trim=0cm 2cm 1cm 2cm,clip]{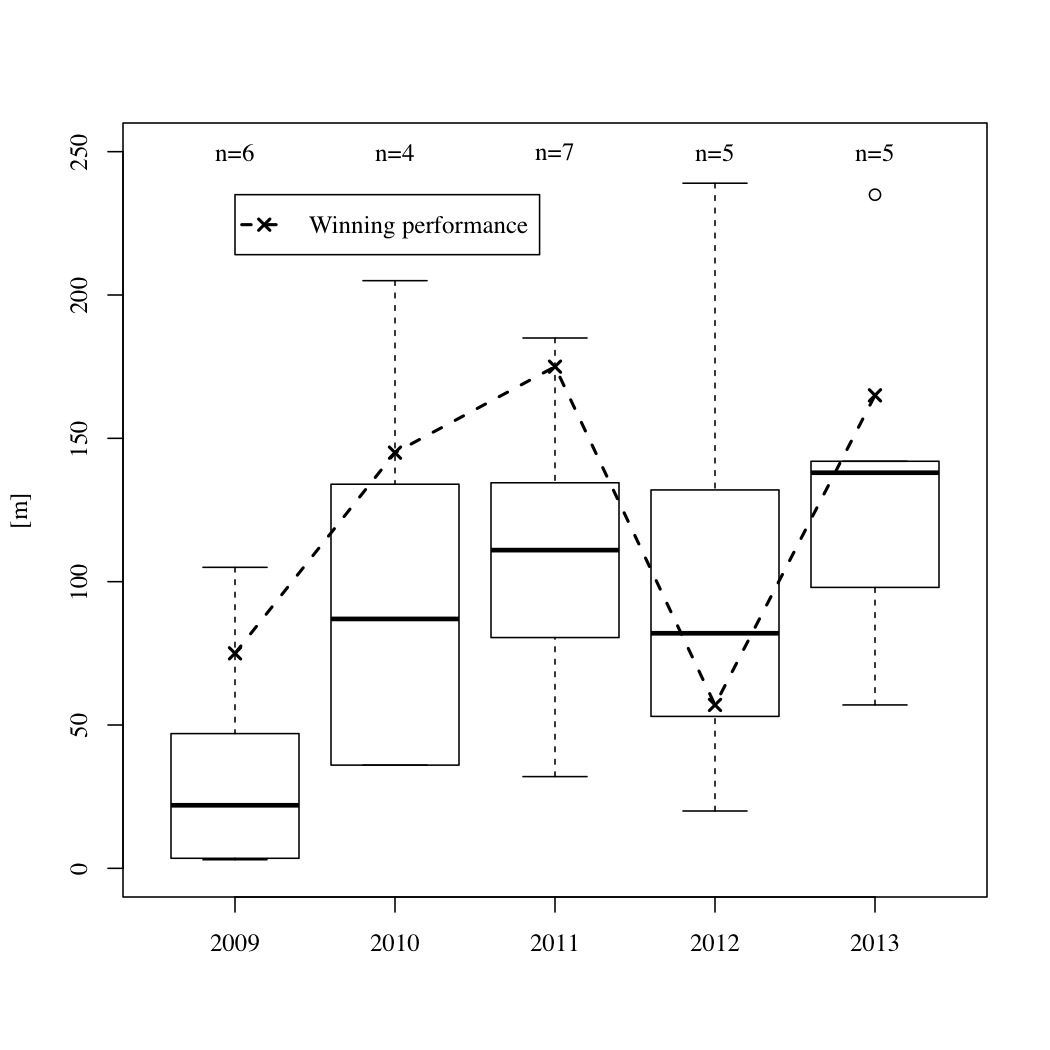}
        \caption{Distribution of driven distances on the track without penalty meters for the task ``overtaking obstacles''.}
        \label{fig:Task2}
  \end{center}
\end{figure}

The second task is to drive as many rounds as possible within 3min on the same
track when obstacles made of white paper boxes with predefined minimum
dimensions are arbitrarily distributed around the track with a focus on curves.
Fig.~\ref{fig:Task2} depicts the distribution of driven distances on the track without
penalty meters. Here, a similar pattern with respect to the median values over
the years can be seen with an average around 140m in the 2013 edition.

\begin{figure}[h!]
  \begin{center}
	\includegraphics[width=.95\linewidth,trim=0cm 2cm 1cm 2cm,clip]{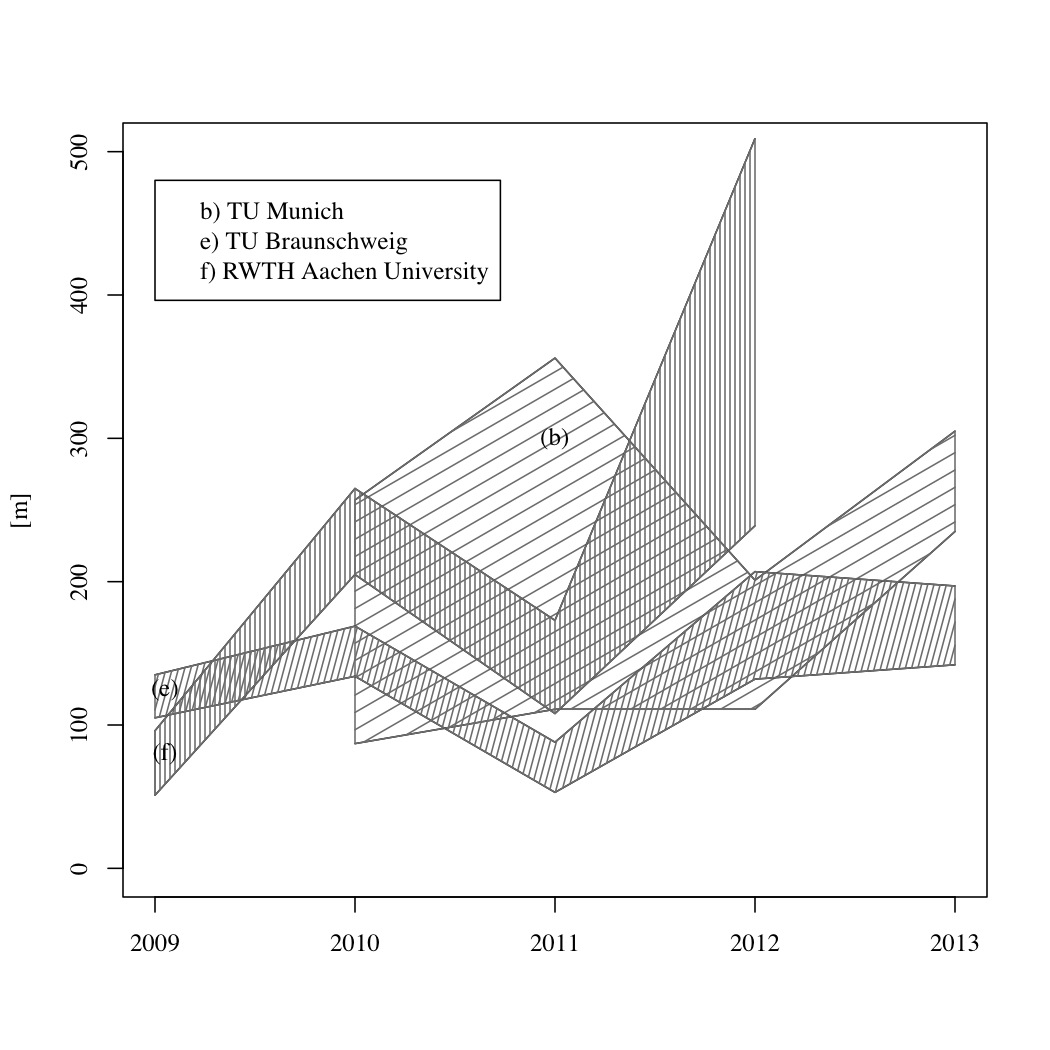}
        \caption{Visualization of driven distances including penalty meters for the teams, who participated at least four times in the last five years.}
        \label{fig:Task2LongTerm}
  \end{center}  
\end{figure}

Fig.~\ref{fig:Task2LongTerm} shows a further analysis of the performance for
those teams who participated at least four times in the last five years; in this
discipline, fewer teams participated on a regular basis. It is obvious that the
difference between totally driven distance and the distance including penalty
meters is pretty constant for the team from TU Braunschweig, in the first three
editions for the team from RWTH Aachen, and in the last two years for the team
from TU Munich. The median velocity for these teams was $0.78\frac{m}{s}$
and for the top team $1.31\frac{m}{s}$ in 2013.

\subsubsection{Task: ``Sideways Parking''}
\label{sec:CaroloCupParking}

\begin{figure}[h!]
  \begin{center}
	\includegraphics[width=.95\linewidth,trim=0cm 2cm 1cm 2cm,clip]{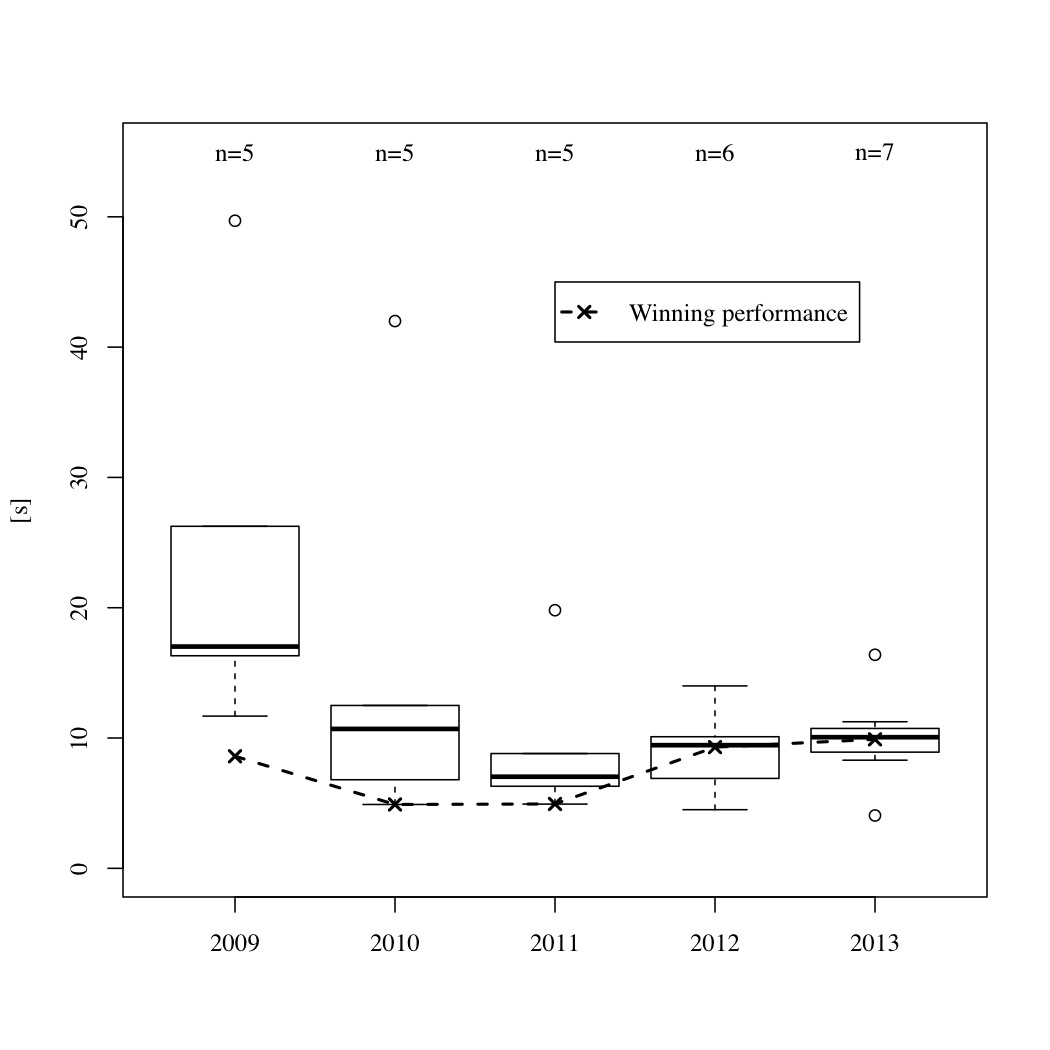}
        \caption{Distribution of the duration for autonomous parking on a sideways parking strip.}
        \label{fig:Task3Average}
  \end{center}  
\end{figure}

The third task requires the teams to park as fast as possible on a sideways
parking strip alongside a straight road without touching other ``parked''
obstacles imitated by white paper boxes. Therefore, the vehicle has to find the
smallest but yet sufficiently wide parking gap among several possible ones.
Fig.~\ref{fig:Task3Average} shows the distribution of the duration for the teams
from the last five years. It can be seen that the teams have on average converged
to a comparable duration for this task in the last four years. For the 2013 edition,
the median duration was $10.74s$ and for the top team $4.39s$.

\vspace{-\baselineskip}

\subsubsection{Concept Classification Matrix}
\label{sec:TechnicalConceptMapping}

After analyzing the results from the last five years of the competition,
technical concepts from the contestants in the 2013 ``CaroloCup'' were
further investigated. Therefore, the individual presentations given by these teams
were analyzed regarding the concepts for environment perception. The findings
were related to the different aspects of self-driving vehicular functions
(alongside the X-axis) and classified according to the different stages for data
processing (alongside the Y-axis) as shown in Fig.~\ref{fig:SensorAlgorithmMapping}.
The following aspects were of interest during the document analysis:

\begin{figure}[htb]
  \begin{center}
	\includegraphics[width=\linewidth]{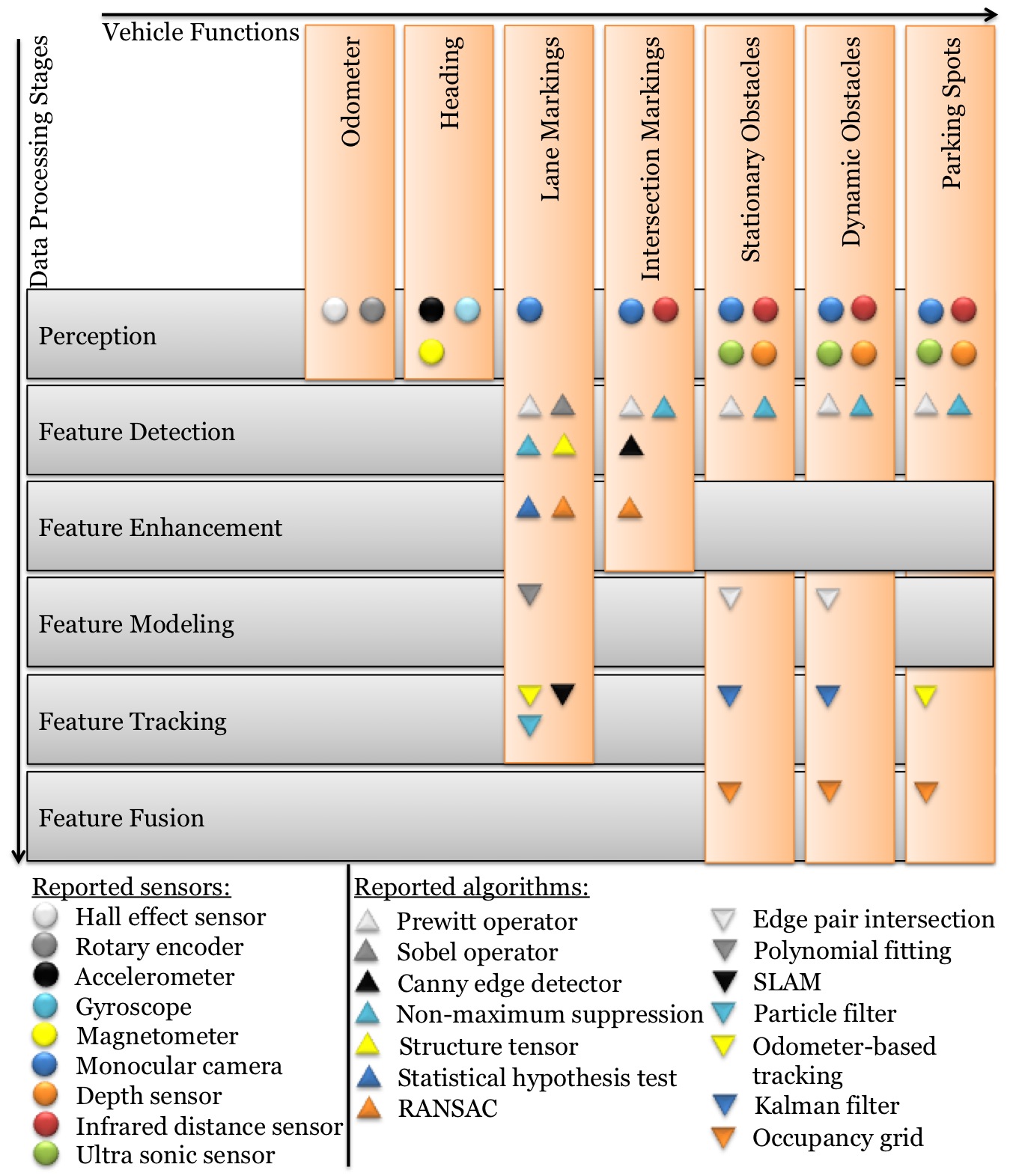}
        \caption{Sensor types and mapping of algorithms to different data processing stages from 2013 CaroloCup participants.}
        \label{fig:SensorAlgorithmMapping}
  \end{center}  
\end{figure}

\begin{itemize}
\item The vertical bars describe different technological aspects for sensing
environmental data for a self-driving vehicle.
\item The horizontal bars describe the different data processing stages.
\item The first horizontal bar \textit{Perception} summarizes the different
sensors, which are reported to be used as the sensor to perceive data for the
given aspect of self-driving vehicular technology. The sensors are denoted by
circles. 
\item The second to last horizontal bars describe different stages in the data
processing, which is supported by various algorithms. The different algorithms
are denoted by triangles.
\item The length of the vertical bars shows up to which data processing stage
the specific aspect of the self-driving vehicular technology is supported by
algorithms.
\end{itemize}

After analyzing the classification matrix, it is apparent that no further data
processing is applied to handle and improve odometer and heading data. This 
could be due to the fact that either no specific algorithms have been reported
or the data quality is sufficiently precise enough to serve properly the use
cases in the competition.

The most effort according to the concepts reported by the teams was spent to
realize reliable and robust lane markings and intersection markings detection: Here, ten
different algorithms at four subsequent data processing stages are applied.
This seems reasonable because having a robust lane-following system, which is
able to also handle missing lane marking segments as required from the rules
and regulations document, is the basis for the other features of the
self-driving car.

Handling obstacles was reported to be supported by six different algorithms and
two different sensor types: Vision-based and distance-based sensors. Here, the
majority of the teams is using ultrasonic and infrared sensors, while some
teams also experiment with depth sensors to retrieve more data from the field
of view. Interestingly, no explicit use of specific algorithms was reported to
increase the feature quality before further processing the data to react on
obstacles in the surroundings. This is in line with the results shown in
Fig.~\ref{fig:Task2LongTerm}, where the competition entries from the long-term
participants show unreliable behavior resulting in 94.29\% additional penalty
meters on average when handling obstacles in the years 2010--2012.

\begin{figure}[t!]
  \begin{center}
	\includegraphics[width=\linewidth,trim=0cm 0cm 0cm 4cm,clip]{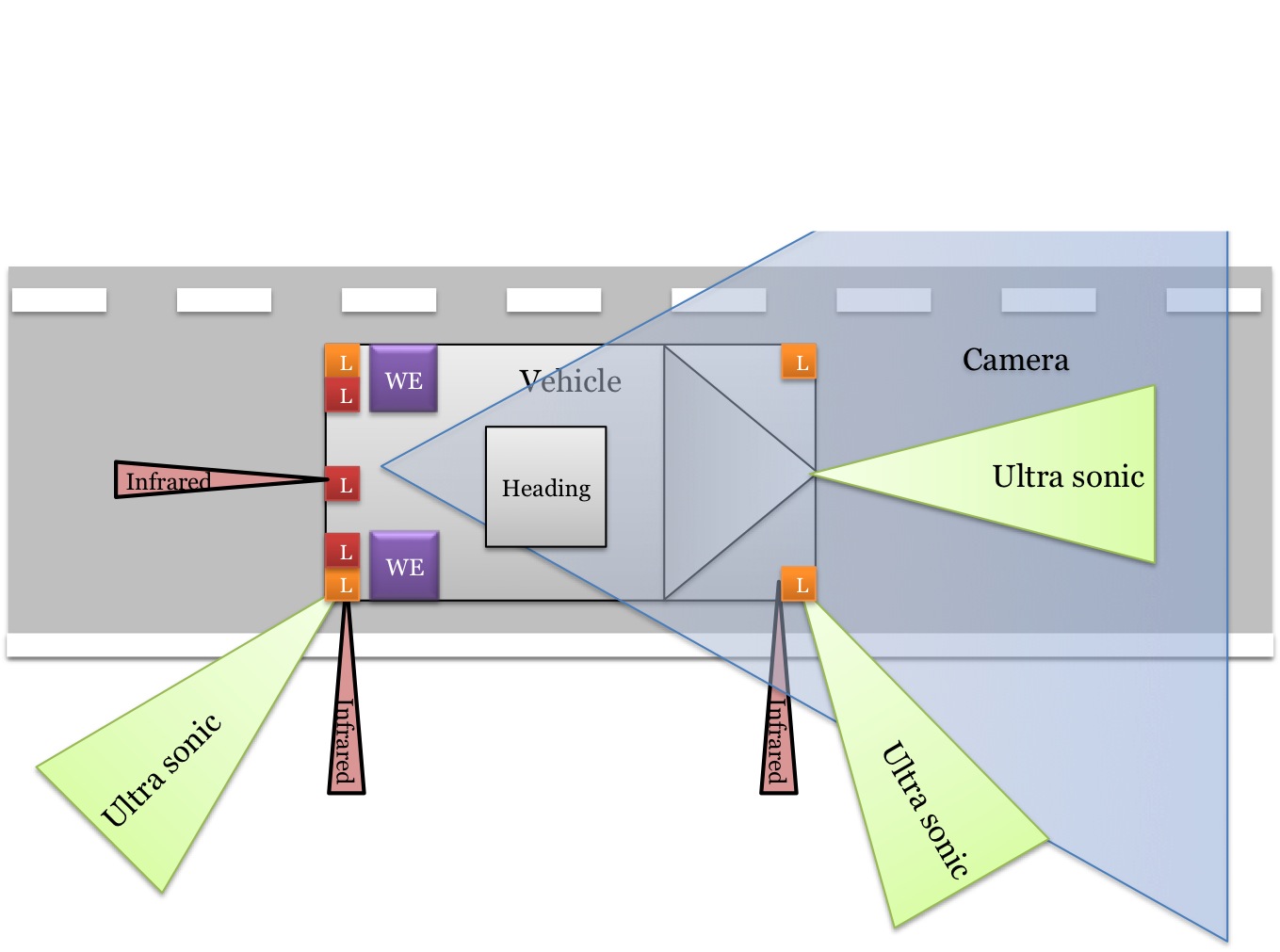}
        \caption{Sensor layout for the required use cases.}
        \label{fig:SensorLayout}
  \end{center}  
\end{figure}

\section{Concepts and Architecture}
\label{sec:Design}

In the following, conceptual considerations for a standardized experimental
platform for self-driving miniature cars are presented and discussed to
address RQ-2 and RQ-3. Firstly, the sensor layout is presented, which
is suitable to address the use cases for a self-driving miniature car as required
by the competition. Next, design drivers for the hardware architecture are
outlined and discussed. Finally, design considerations for the software
architecture of the platform are discussed where a specific focus is given to
simulation-based evaluation.

The findings that are presented in this section are based on several sources:
(a) Results from the SLR presented before, (b) conclusions drawn from the
systematic analysis of the results from the last years' competitions, (c)
findings from the concept classification matrix, and finally, (d) experiences
from the development of the self-driving miniature car ``Meili'', which won
the ``Junior Edition'' in the 2013 competition.

\subsection{Sensor Layout}

As it is evident from Fig.~\ref{fig:SensorAlgorithmMapping}, the dominant
sensors are distance sensors like ultrasonic and infrared to detect obstacles, which
is reported in 50\% and respectively 75\% of the team presentations, and
vision-based sensors to perceive lane markings as reported in all presentations.
Furthermore, 75\% of the teams reported in their technical presentations the use
of incremental sensors to measure the travelled path.

Fig.~\ref{fig:SensorLayout} depicts the resulting sensor layout, which was derived
from the required use cases for the competition: Wheel encoders (labelled WE in
the figure) are mounted at the rear axle to measure the driven distance and a
monocular camera is used to perceive information about the lane markings.
The ultrasonic- and infrared-based distance sensors are focusing on the right
hand side (a) to serve the parking and (b) to handle overtaking situations.
Furthermore, the rules and regulations document states that right-of-way
situations will occur where the self-driving vehicle has to yield the right-of-way
to another vehicle. Thus, no focus is given to obstacles on the left hand side
of the vehicle. Both distance sensors located at the vehicle's rear-end enable
dynamic adjustments of the vehicle in the parking spot.

\begin{figure}[t!]
  \begin{center}
	\includegraphics[width=\linewidth]{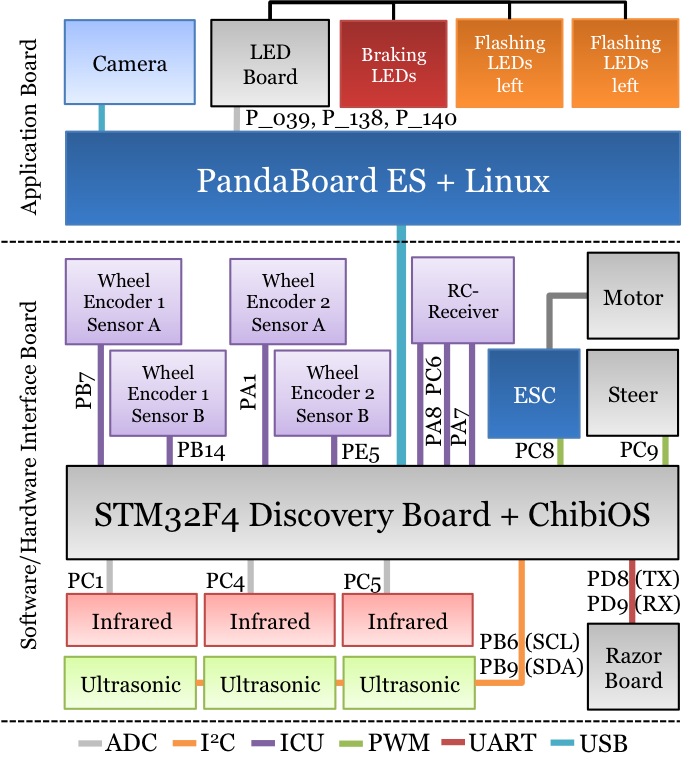}
        \caption{Hardware architecture and pin connection plan.}
        \label{fig:HardwareArchitecture}
  \end{center}  
\end{figure}

The experience of the development of ``Meili'' also showed that using
ultrasonic- and infrared-based distance sensors is beneficial due to their low
power consumption and their low costs in comparison to depth camera sensors
or small-scale laser scanners. Furthermore, due to their simple technical
interfaces, both types of distance sensors are easy to integrate and to maintain.

The competition takes place indoors and thus, the use of GPS is not recommended.
Alternatively, an anchor-based system as suggested by Pahlavan et al.~\cite{PPS11}
could be used to realize a precise indoor localization. While such a system would
not be allowed to be used during the competition, it could serve as a reference
system for evaluating algorithms on the experimental platform on the test track.

\subsection{Design Drivers for the Hardware Architecture}

The software/hardware interface board to perceive data from the sensors or to
control the actuators for ``Meili'' in the 2013 competition was designed and
realized as described by Vedder \cite{Ved12}. That design relied on a
self-assembled printed circuit board (PCB), which allowed a very compact and
resource-efficient realization of the hardware interface. However, the main
drawback for that crucial component was the time-consuming and error-prone
manufacturing process that could delay the entire integration and testing
of all components later. Furthermore, maintaining the software on the PCB
required a specific hardware and software flashing environment, which added
further complexity to the development and maintenance process.

Based on this experience, important design drivers for the hardware architecture
are availability and reusability of components especially when parts are broken
while having a strong focus on low effort integration. Therefore and as preliminarily
outlined in \cite{BMH13}, using commercial-of-the-shelf (COTS) components
enabled these requirements especially in terms of quality and costs.

\begin{figure}[t!]
  \begin{center}
	\includegraphics[width=.9\linewidth,trim=0cm 0cm 0cm 0cm,clip]{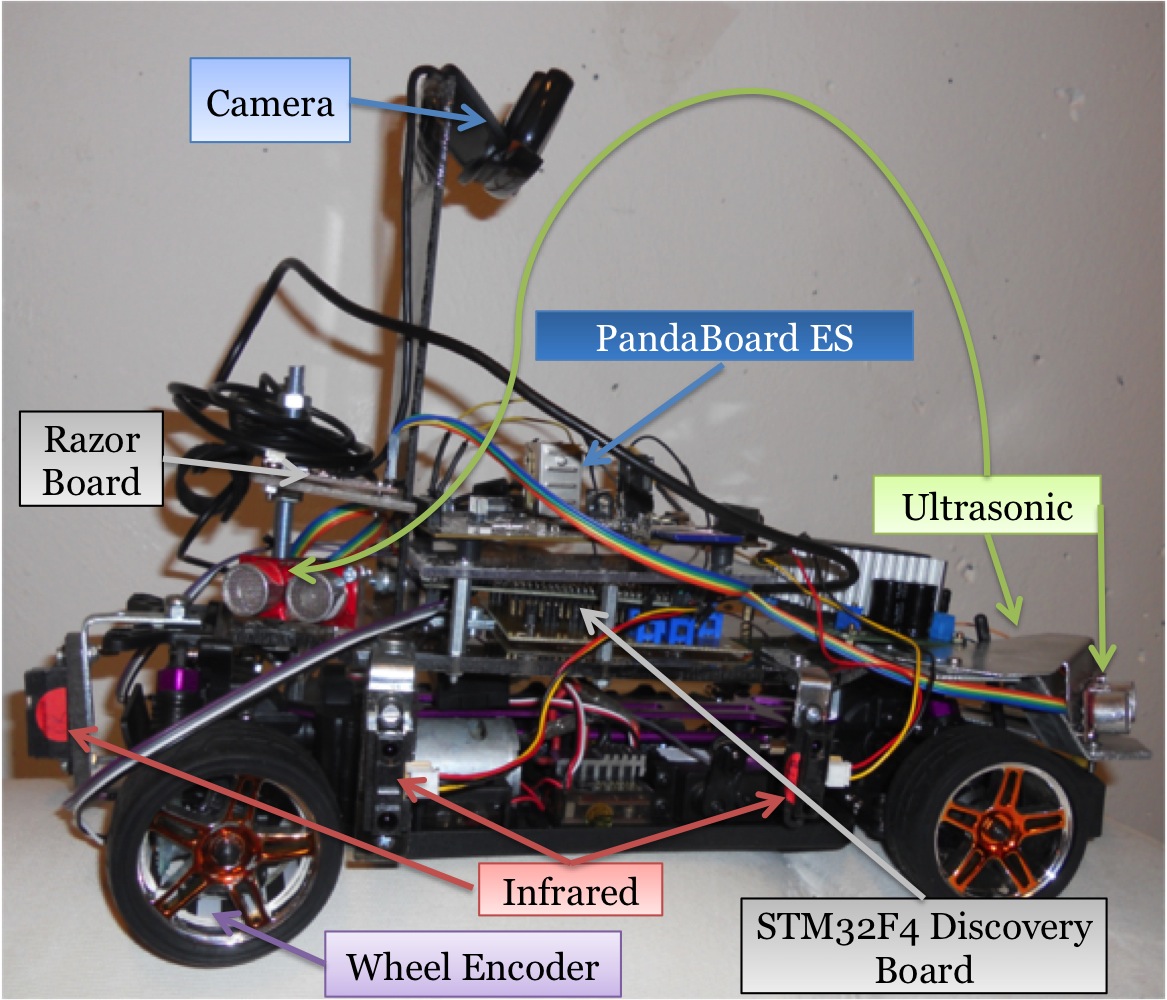}
        \caption{1/10 scale vehicle realizing the hardware concept.}
        \label{fig:Car}
  \end{center}  
\end{figure}

The hardware architecture is separated into two parts: (a) Application board
and (b) software/hardware interface board, as shown in
Fig.~\ref{fig:HardwareArchitecture}. The ARM-based application board runs
a Debian-based Linux with all software components to realize the self-driving
vehicle functionality: Image capturing and feature detection, lane-following,
overtaking, and sideways parking. The only sensor that is directly connected to
this board is the monocular camera due to the required computation power to
handle its input data.

Embracing the experiences from the development of ``Meili'', the
software/hardware interface board is realized with the ARM-based STM32F4
Discovery Board\footnote{www.st.com}, which provides enough pins to serve
the required sensors and actuators as described by the sensor layout in
Fig.~\ref{fig:SensorLayout}. This board interfaces with the ultrasonic- and
infrared-based distance sensors, the wheel encoders, the inertial measurement
unit to determine the vehicle's heading by evaluating accelerometer and gyroscope
data, the emergency override for the RC-handset, and the steering servo and
acceleration motor. The board itself is connected via a serial-over-USB
connection to the application board to enable bi-directional communication.

Furthermore, the STM32F4 Discovery Board allows the extension of the sensor
layout as depicted in Fig.~\ref{fig:SensorLayout} by up to five more infrared-based
and 13 further ultrasonic-based distance sensors for future use cases and
hence, having 24 distance sensors in total. An example of a miniature vehicle
realizing the outlined concept is depicted by Fig.~\ref{fig:Car}.

\subsection{Design Drivers for the Software Architecture}

The design decision for COTS components enabled reusability and flexibility
on the hardware architecture; however, this decision also bears the risk that
the selected components are not available anymore over time and hence, other
boards with different hardware configurations need to be used. Thus, the
software architecture needs to compensate for this flexibility in the hardware
architecture.

\subsubsection{Software/Hardware Interface}

This compensation is reflected as reusability of the components from the software
architecture, which interface with the sensors, actuators, or rely on system calls
of the underlying execution platform. Therefore, a hardware abstraction layer
(HAL) is required to reduce these dependencies on the real hardware. For the
software/hardware interface board, the open source real-time operating
system (OS) ChibiOS/RT\footnote{www.chibios.org} was used on the STM32F4
Discovery Board to have a standardized hardware abstraction layer as
programming interface. This OS supports several different embedded systems
and thus, the low-level software to interface with sensors and actuators can be
easily reused in future robotics projects as well. Furthermore, ChibiOS/RT is well
documented and the resulting binary exhibits a small memory footprint.

Having a standardized software/hardware interface also enables generative
software engineering like model-checking and code generation as outlined
in \cite{MBH13}. In this regard, the software components that are interfacing
with the sensors and actuators can be adjusted automatically when more sensors
are added or the configuration is modified.

\subsubsection{Application Environment}

The self-driving functionality is composed by software components that realize
image processing and lane-following with overtaking, as well as sideways parking.
The components are designed according to the \textit{pipes-and-filters}
principle \cite{GHJV94} running at a constant frequency while reading their
input from data sources like a camera, distance sensors, or odometer to
compute the required set values as described in Sec.~\ref{sec:LaneFollowing}
and Sec.~\ref{sec:Parking} to control the vehicle.

To complement the aforementioned embedded real-time OS, which is
used as HAL for the software/hardware interface, the portable middleware as
described in \cite{Ber10} was used. This middleware written in ANSI C++
provides a component-based execution and data processing environment with
real-time capabilities enabled by Linux rt-preempt \cite{MSR07}.

The interacting components running on the application board are processing
platform-independent data structures, which represent generic distances for
the ultrasonic and infrared sensors as well as the information about vehicle
heading and travelled path, and generic image data. Thus, the self-driving
functionality is standardized in such a way that it is not depending on
elements of a concrete hardware environment. Instead, these data structures
are mapped to a platform-specific realization by an additional component
that realizes the structural design pattern \textit{proxy} \cite{GHJV94}, which
connects the software components from the application environment with
the embedded software running on the software/hardware interface board. 

This design decision based on experiences from a successful realization of
several real-scale self-driving cars \cite{RBL+08,Ber10}. On the one hand,
it supports the standardization of the self-driving functionality by reducing
dependencies on concrete components of the hardware environment, and on
the other hand, it allows seamless reusability of the software components from
the application environment in a virtual test environment for evaluating
algorithmic concepts before their deployment on the experimental hardware
platform.

\subsection{Virtual Test Environment}

Having a virtual test environment available to conceptualize and evaluate
algorithms has been proven a successful approach for real-scale
self-driving cars \cite{BR12a}. According to the technical concept
classification matrix in Fig.~\ref{fig:SensorAlgorithmMapping}, the simulation
system, which realizes the virtual test environment, needs to include models
to produce images from the virtual surroundings of the vehicle, and models
for producing data for distance-based sensors. Furthermore, the simulation
needs to include a physical motion model for the vehicle dynamics to enable
a closed-loop experimentation system, where algorithms process data from
virtualized sensors to act in response to the surroundings.

There are several simulation environments for robotic platforms available like
GAZEBO \cite{GVH03,HKH12}, which was successfully used during the DARPA
Robotics Challenge, MORSE \cite{ELD+12}, or ARGoS \cite{PTG12}. These
platforms mainly focus on general experimental robotic platforms and hence,
domain-specific modeling support to describe large scale and different
automotive scenarios with roads, intersections, and parking lots to be used
for validating algorithms for automotive functions is missing.

Furthermore, the aforementioned simulation environments can be integrated
with the middleware ROS \cite{QGC+09}, which allows a
convenient realization of communication between distributed processes
by encapsulating the low-level and OS-specific implementation details; a
similar approach is also realized by the lightweight communications and
marshalling (LCM) middleware \cite{HOM10}. While both software environments
provide a well structured message interchange, they do not support means to
fully control and schedule the communication in case of coordinated simulations.

In contrast, the concepts and middleware used on the self-driving miniature
car ``Meili'' were already successfully used on real-scale self-driving
cars \cite{RBL+08,Ber10} while explicitly focusing on seamless integration with
a domain-specific simulation environment. While the middleware also realizes
a transparent communication based on UDP-multicast, the interacting components
can be fully controlled and transparently interrupted when they are embedded in
a simulation environment. Thus, synthetic data from virtualized sensors can
seamlessly replace real sensor data to validate algorithmic concepts.

The simulation environment that was used for the self-driving miniature car
(a) provides a formalized model of the environment by using a domain-specific
language (DSL), (b) has virtualized counterparts for the sensors of interest,
and (c) provides a model for the vehicle dynamics. To reuse the existing
environment, which was used so far for real-scale vehicles, the 1/10 scale
environment and its constraints for the miniature vehicles were scaled up.

\begin{figure}[h!]
  \begin{center}
	\includegraphics[width=\linewidth]{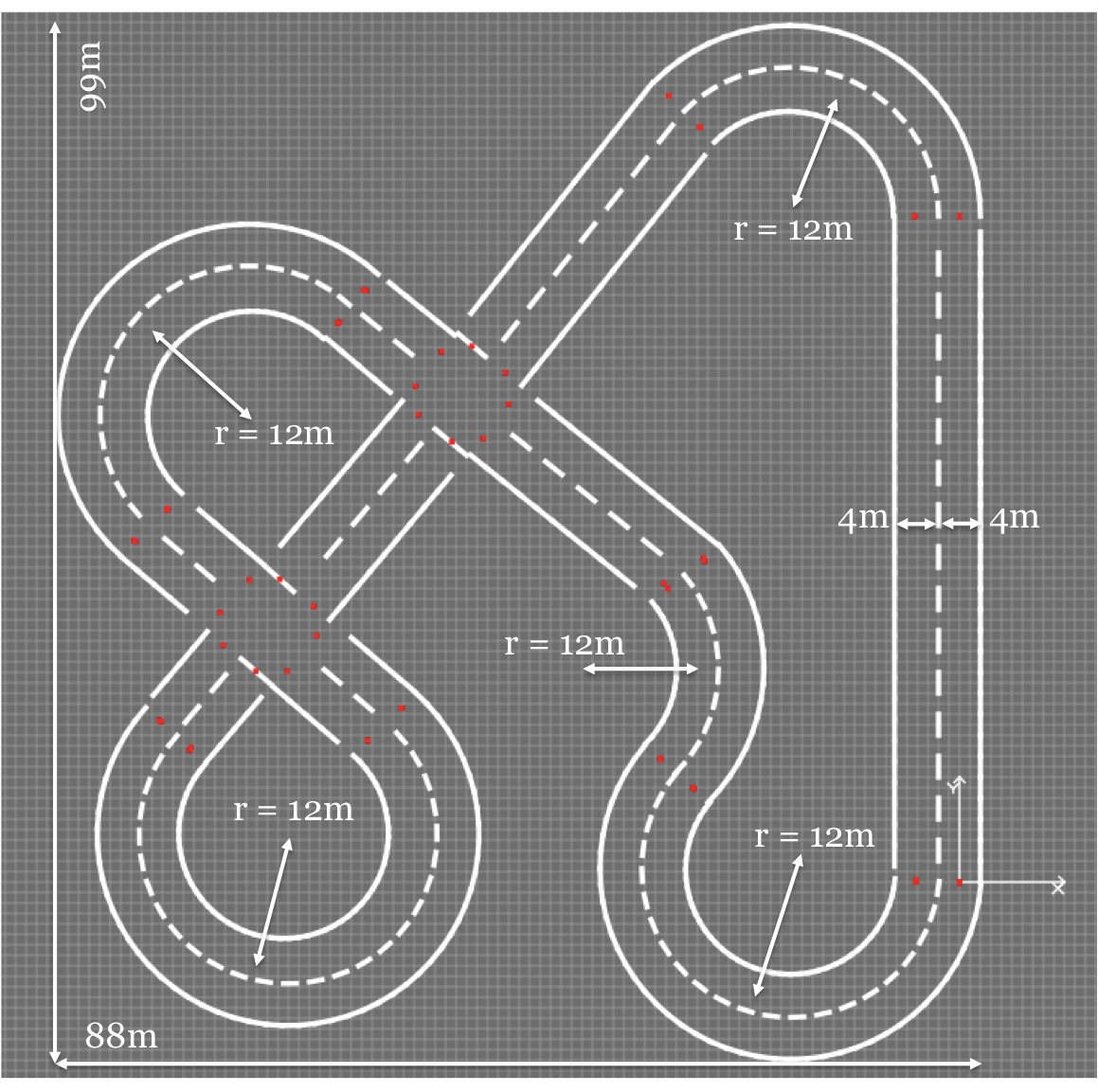}
        \caption{Model of the environment in the simulation.}
        \label{fig:TrackAnnotations}
  \end{center}  
\end{figure}

To create an appropriate model for the environment, the rules and regulations
document of the competition was used as a basis because it defines minimum
dimensions for roads, lanes, curvatures, and intersections. Furthermore, it
provides a potential setup of a track following these constraints in its appendix.
The same layout was modeled while considering the aforementioned constraints.
Since the minimum radius of the track was defined as 1m measured up to the
inner side of the lane marking from the outer lane and considering a lane's width
as 0.4m, this radius is scaled up to 12m up to an inner curve's skeleton line. The
visualized 2D variant of the resulting model for the exemplary track is shown in Fig.~\ref{fig:TrackAnnotations}.

The technical concept classification matrix in Fig.~\ref{fig:SensorAlgorithmMapping}
as well as the analysis of the technical presentations of the teams unveiled
monocular cameras as well as distance sensors like ultrasonic- and infrared-based
sensors as predominantly chosen sensors. Thus, these sensors needed to be
virtualized in the simulation environment as well.

For the former image-providing sensor, a DSL's instance is also transformed to
a 3D-OpenGL-representation of the surroundings. This representation is then
used to capture perspectively correct virtualized monocular camera images
depending on the virtualized mounting position of the camera and the vehicle
position. This image sequence can subsequently be used to realize the
fundamental vehicle function lane-following. A concept and an evaluation for a
lane-following control algorithm based on such image data is described in
Sec.~\ref{sec:LaneFollowing}.

The distance sensors like ultrasonic and infrared sensors were added to the
existing simulation environment. Therefore, their virtualized mounting positions
were modeled, alongside with their opening angle and up-scaled viewing distance.
Depending on the vehicle's current position, a single-layer, ray-based algorithm
determines the set $I$ of intersection points per sensor with the modeled
surroundings like obstacles or parked vehicles and returns the nearest distance
$d$ to a sensor's mounting position.

The vehicle motion model was reused from the existing simulation environment
basing on the bicycle-model \cite{Ber10}. Furthermore, the distance between the
front and rear axle, and the minimum turning radii to the left and right hand side
were determined and scaled up to parametrize the vehicle model from the
real-scale environment.

This virtual test environment was used to design, realize, and evaluate the
algorithmic concepts that are required for the competition: Firstly, an approach
for an overtaking algorithm as part of a lane-following control algorithm is
described in Sec.~\ref{sec:LaneFollowing}, and one for a parking algorithm
based on distance-sensors is outlined in Sec.~\ref{sec:Parking}.

\section{Lane-Following with Overtaking}
\label{sec:LaneFollowing}

In the following, the general design considerations for a state-machine to realize the
fundamental behavior of following lane markings and overtaking obstacles based
on image data is described to address \textit{RQ-4}. The focus for this section is
to outline the basic ideas behind a lane-following concept and its evaluation in
the simulation environment.

\subsection{Design Considerations}

\subsubsection{Lane-Following Algorithm}

The idea behind the fundamental lane-following algorithm is based on the
calculation of the vehicle's deviation from a lane's skeleton line. This concept
does not necessarily result in the shortest or fastest path through a given road
network since better trajectories could be found \cite{Kar08}; however, it forms
the basis for more advanced algorithms that, e.g., incorporate self-localization
and mapping (SLAM) \cite{DNCD+01,LMT08,LT10} to extend the basic concept
presented here.

\begin{figure}[t!]
  \begin{center}
	\includegraphics[width=\linewidth,trim=1cm 0cm 1cm 0cm,clip]{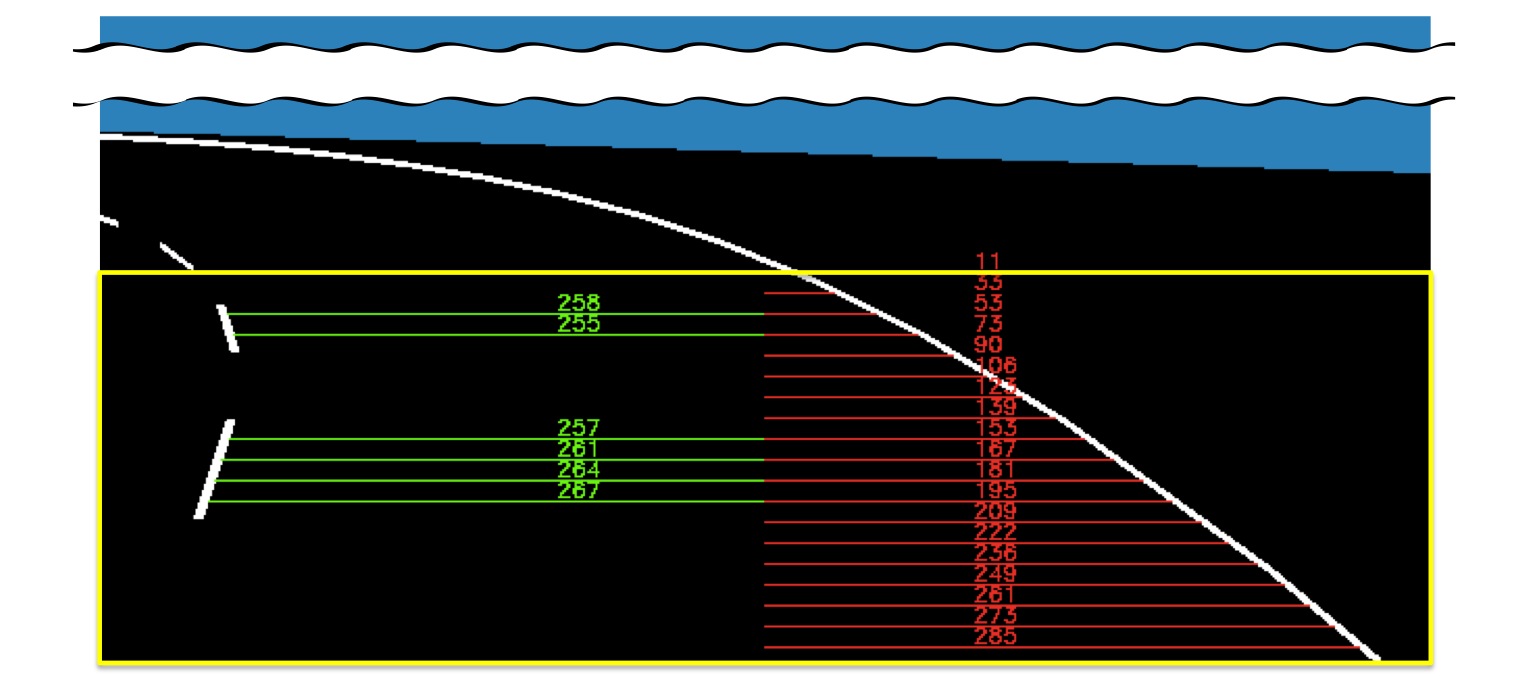}
        \caption{Image feature detection based on virtualized input images from the simulation environment with shortened horizon and a yellow region-of-interest in the lower-level 40\% of the image. The average processing time per frame is $0.719ms$ on a 1.8GHz Intel Core i7 with 4GB RAM.}
        \label{fig:LaneFollowingAlgorithm}
  \end{center}
\end{figure}

Fig.~\ref{fig:LaneFollowingAlgorithm} depicts the basic principle behind the
algorithm. Firstly, it is assumed that the vehicle is positioned between two
lane markings heading towards the desired driving direction. Secondly, several
scan-lines starting from the image's bottom are used to calculate the deviation
from the vertical center line in the image to the left and right hand side until
a white lane marking is found. To calculate the desired steering angle, the
actually measured distance per scan-line from one side is compared either
(a) to a previously calibrated desired distance resulting in the desired
steering angle to follow this lane marking, or (b) to the distance to the
other hand side with the goal to have an equalized distance to the left and
right hand side per scan-line, which results in a desired steering angle.
\begin{figure}[t!]
  \begin{center}
	\includegraphics[width=.95\linewidth,trim=0cm 0.5cm 1cm 2cm,clip]{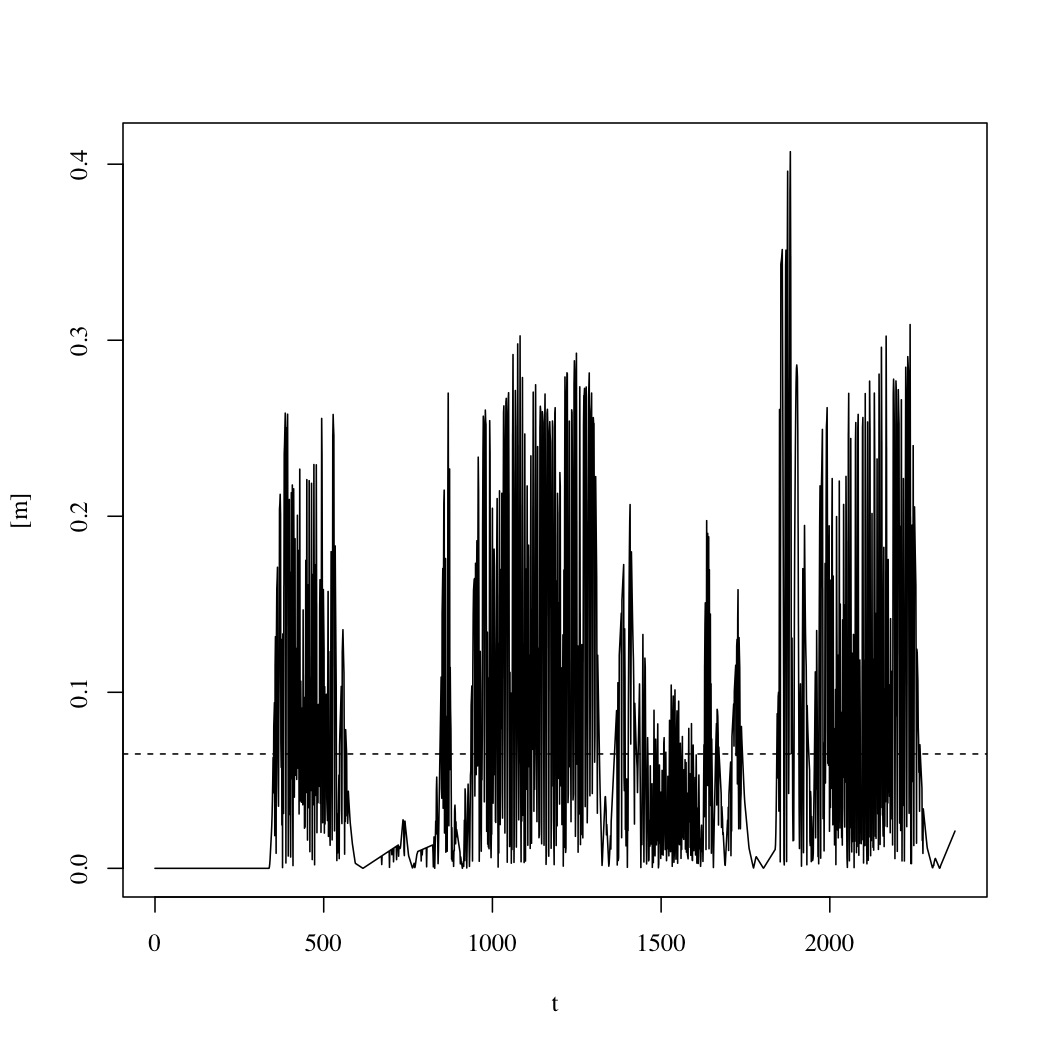}
        \caption{Deviation from a lane's skeleton line is $\bar{d} = 0.065m$. The highest peak happens during passing an intersection.}
        \label{fig:Deviation}
  \end{center}
\end{figure}

\begin{figure*}[t!]
  \begin{center}
	\subfigure[Data from several distance sensors over time in the simulation environment.]
	{
		\includegraphics[width=.475\linewidth]{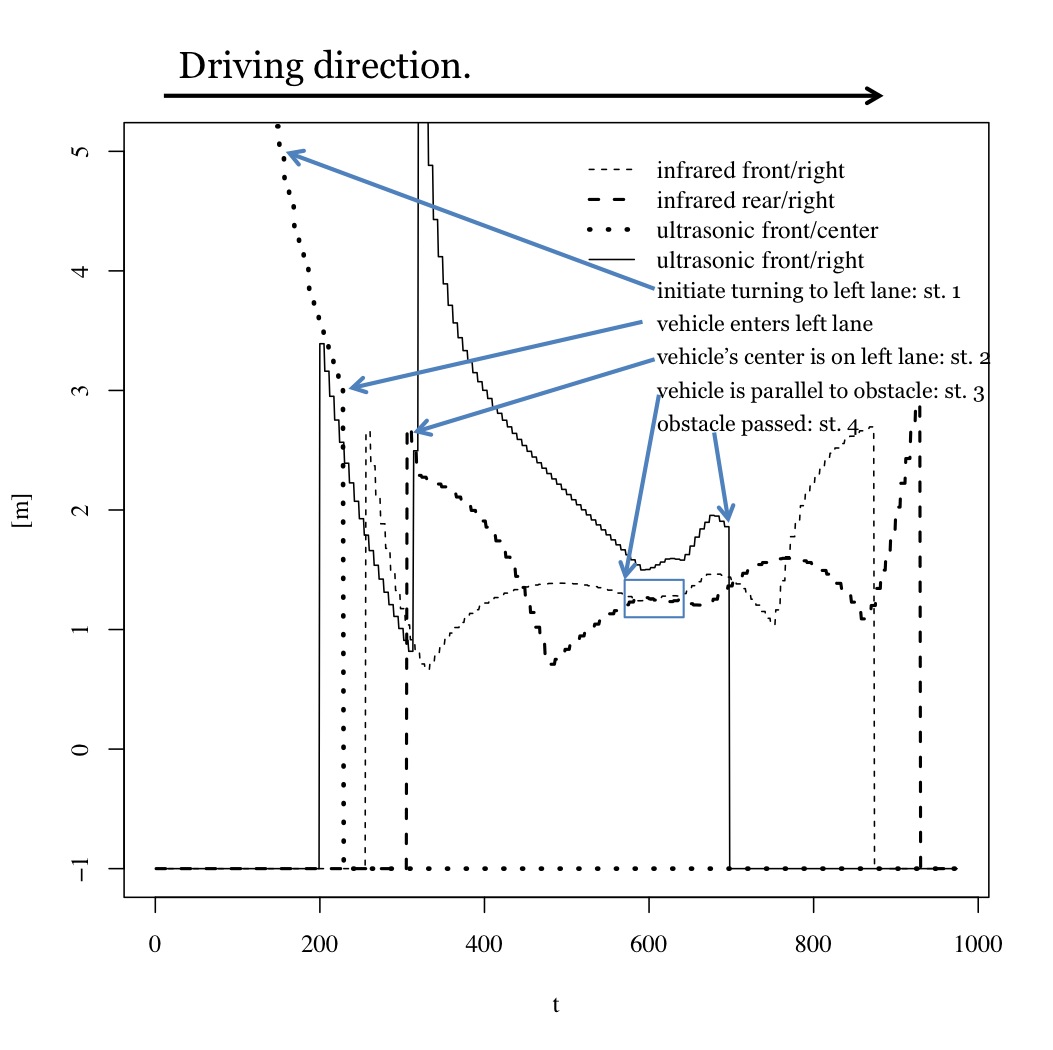}
	        \label{fig:DistancesOvertakingAnnotations}
	}
	\subfigure[Overtaking scenario in the simulation environment.]
	{
		\includegraphics[width=.475\linewidth]{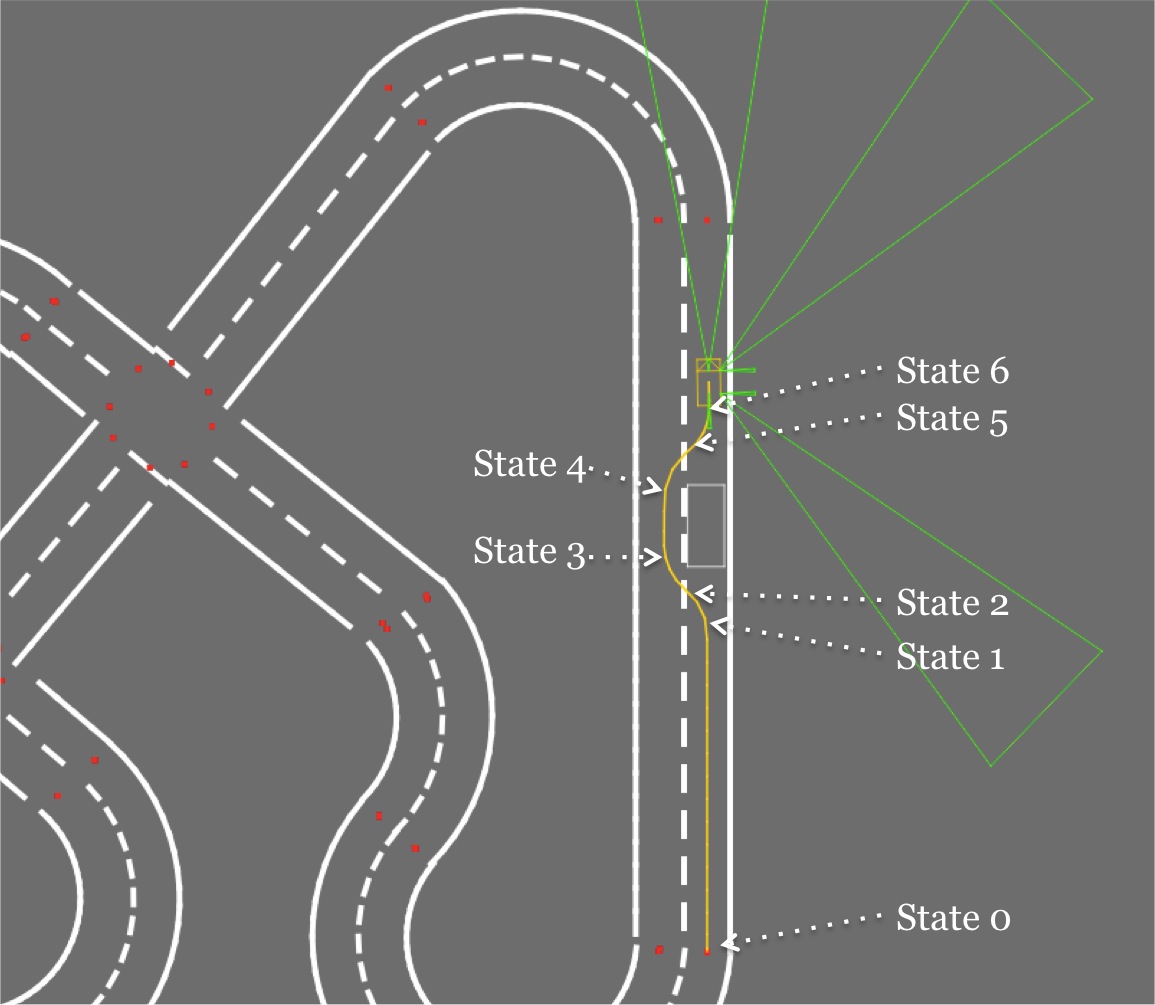}
        	\label{fig:ObstaclePassed}
	}
        \caption{Concept for the overtaking algorithm.}
        \label{fig:Overtaking}
  \end{center}  
\end{figure*}

Another possibility utilizes several scan-lines starting from the image's
bottom towards the image's center. Hereby, the deviation of the center points
per scan-line from the vertical image's center line is calculated resulting in
a point sequence, which can be fitted by the best matching arc having its
center point on the left or right hand side of the vehicle depending whether it
is a left or right curve. This arc in turn represents the curvature that is
needed to follow the lane ahead of the vehicle. The advantage of the arc-fitting
algorithm is that it can handle missing lane markings more robustly since missing
or not plausible center points, which are not following the fitted lane-model,
can simply be omitted \cite{LSBL+08}.

Fig.~\ref{fig:Deviation} shows an evaluation of the lane-following algorithm using
the following PI-controller based on the first approach:
$y(t) = 2.5 \cdot e + 8.5 \cdot \int e(\tau) d\tau$, where e is the distance error
for the scan-line of interest. On average, the car's deviation from a lane's
skeleton is about 6.5cm with peaks up to 30cm in intersection areas, where
lane-markings are missing.

\subsubsection{Overtaking Algorithm}

\begin{figure*}[htb]
  \begin{center}
	\includegraphics[width=\linewidth]{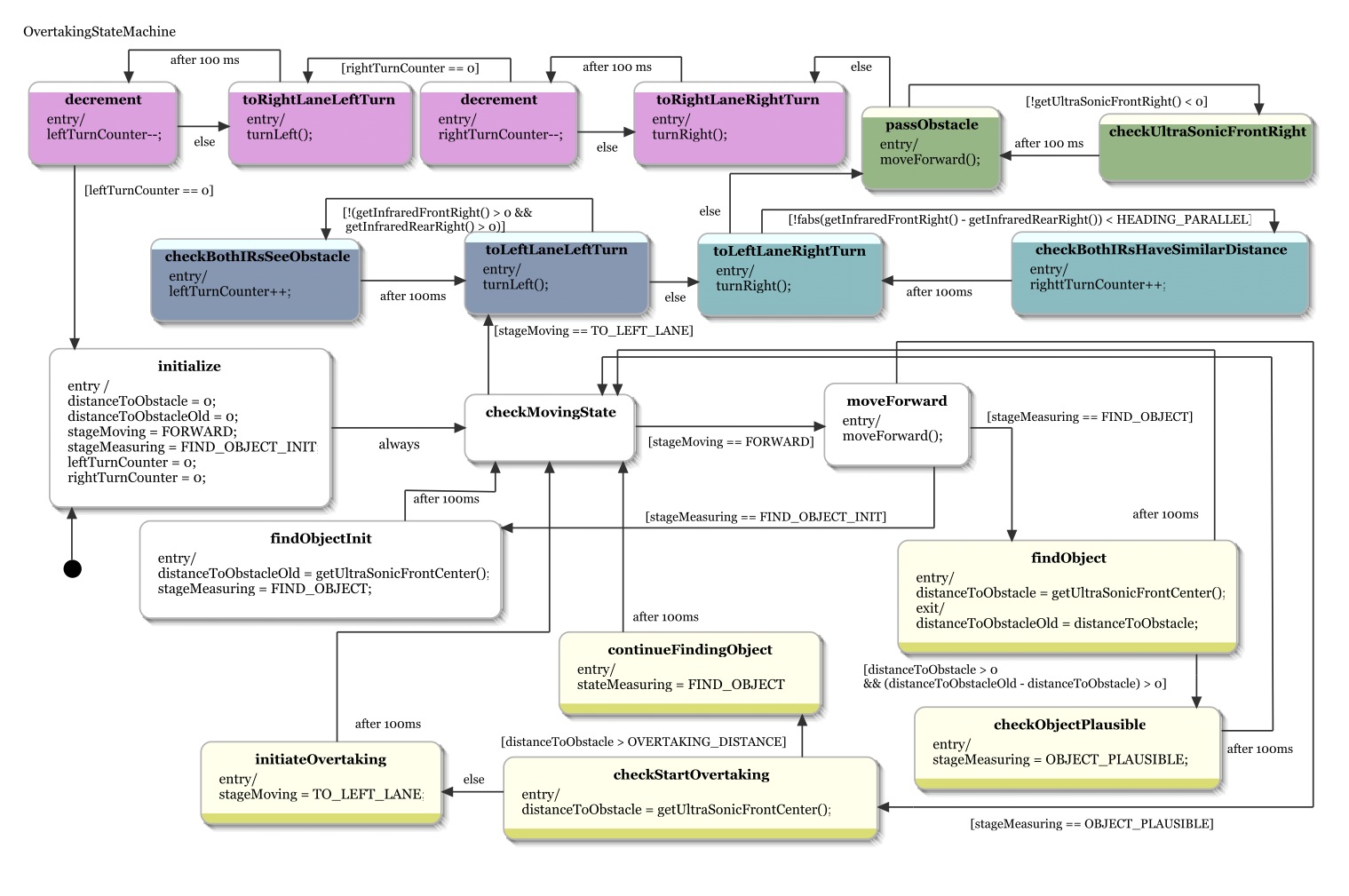}
        \caption{Overtaking state machine.}
        \label{fig:OvertakingStateMachine}
  \end{center}  
\end{figure*}

The concept for the overtaking algorithm using the distance sensors is depicted
in Fig.~\ref{fig:Overtaking}. The basic idea behind it is to use the appropriate
sensors for the different stages of an overtaking process as annotated in 
Fig.~\ref{fig:ObstaclePassed} from the simulation: (a) the vehicle approaches
an obstacle blocking its own lane. This fact is determined by the data perceived
by the ultrasonic front sensor as depicted in
Fig.~\ref{fig:DistancesOvertakingAnnotations}. Once the values have fallen below
a given threshold, the vehicle's trajectory is modified so that it starts
moving to the neighboring lane by steering at maximum to the left. This part of
the lane changing trajectory is terminated at time point (b) once both
infrared-based distance sensors mounted at the right hand side have ``seen''
the obstacle for the first time. The current state of the vehicle according to 
Fig.~\ref{fig:DistancesOvertakingAnnotations} and \ref{fig:ObstaclePassed} is
now interpreted as such that the vehicle's center is on the left lane.

Now, the vehicle needs to steer with a maximum to the right to orientate it in
parallel to the obstacle. This part of the trajectory is terminated at time
point (c) once both infrared-based distance sensors return the same distance
to the obstacle. Next, the vehicle continues on the neighboring lane to actually
pass the obstacle until the ultrasonic distance sensor mounted at the top/right
corner does not ``see'' the obstacle anymore at time point (d).

Finally, the vehicle returns to its original lane by driving the inverted
trajectory from the beginning of the overtaking process. Therefore, the
algorithm has tracked the duration during the left and right arc in the initial
overtaking phase.

\subsection{Lane-Following and Overtaking State-Machine}

Fig.~\ref{fig:OvertakingStateMachine} depicts the overall state-machine to
realize the aforementioned algorithmic concept, which is executed at 10Hz.
The main design goal for the state-machine is the separation of the actual
overtaking process from the lane-following algorithm. The reason for this
design driver is to allow improvements and extensions to the lane-following
algorithm while preserving the overtaking capabilities and vice versa.

The architecture of the state-machine is divided into two main parts:
(a) observing the front area of the vehicle to determine when to initiate the
overtaking process, and (b) the five parts of the overtaking trajectory shown
in the upper half of Fig.~\ref{fig:OvertakingStateMachine}. The first part of
the state machine begins with continuing the actual lane-following process,
which is realized by the state \texttt{moveForward}. The current implementation
uses in this case the acceleration and steering set values as determined from
the image processing and lane-following algorithm without further manipulation.

Subsequently, it validates measurements from the ultrasonic front sensor.
Therefore, only those obstacles are considered, which are either stationary or
driving slower in the same direction as the self-driving miniature vehicle as
realized by the transition to state \texttt{checkObjectPlausible}. Once a
plausible object has been found, its distance to the self-driving miniature
vehicle is validated to initiate the actual overtaking by changing to state
\texttt{toLeftLaneLeftTurn}.

The purpose of the states shown in dark blue in Fig.~\ref{fig:OvertakingStateMachine}
is to steer at maximum to the left until both infrared sensors mounted at the
right hand side of the vehicle have ``seen'' the obstacle. Due to their mounting
position, the center of self-driving miniature vehicle has now entered the
neighboring lane. During this part of the state-machine, a counter is
incremented to record the duration of the left turning part for the lane change
and the regular lane-following algorithm is deactivated.

Afterwards, the vehicle steers at maximum to the right until both infrared
sensors mounted at the right hand side return the same distance with respect to
a given threshold. Once these distances are measured, the self-driving miniature
vehicle is now oriented in parallel to the obstacle and the right turning part
for the lane change can be terminated. A second counter is incremented according
to the duration for this part as well.

The third part in the overtaking state-machine consists of the actual obstacle
passing. This part is highlighted in green in the top right corner of
Fig.~\ref{fig:OvertakingStateMachine}. Here in state \texttt{passObstacle},
the modular lane-following algorithm is activated again and used to follow the
lane markings on the neighboring lane until the ultrasonic sensor mounted at
the vehicle's front right corner does not ``see'' the obstacle anymore.

On this event, the last two parts of the overtaking algorithm take place. Their
purpose is the return to the original lane. Therefore, the lane-following
algorithm is deactivated again and the vehicle steers at maximum to the right
until the second counter representing the second part of the lane changing
process has reached zero. Afterwards, the vehicle steers to the left again
to orient its heading in the correct driving direction again while decrementing
the first counter. Finally, the actual lane-following is activated to continue
on the original lane again and the state-machine is reset to handle the next
obstacle.

\section{Sideways Parking}
\label{sec:Parking}

\begin{figure*}[htb]
  \begin{center}
	\subfigure[Data from ultrasonic front/right distance sensor over time in the simulation environment.]
	{
		\includegraphics[width=.4\linewidth,trim=1cm .5cm 1cm 1cm]{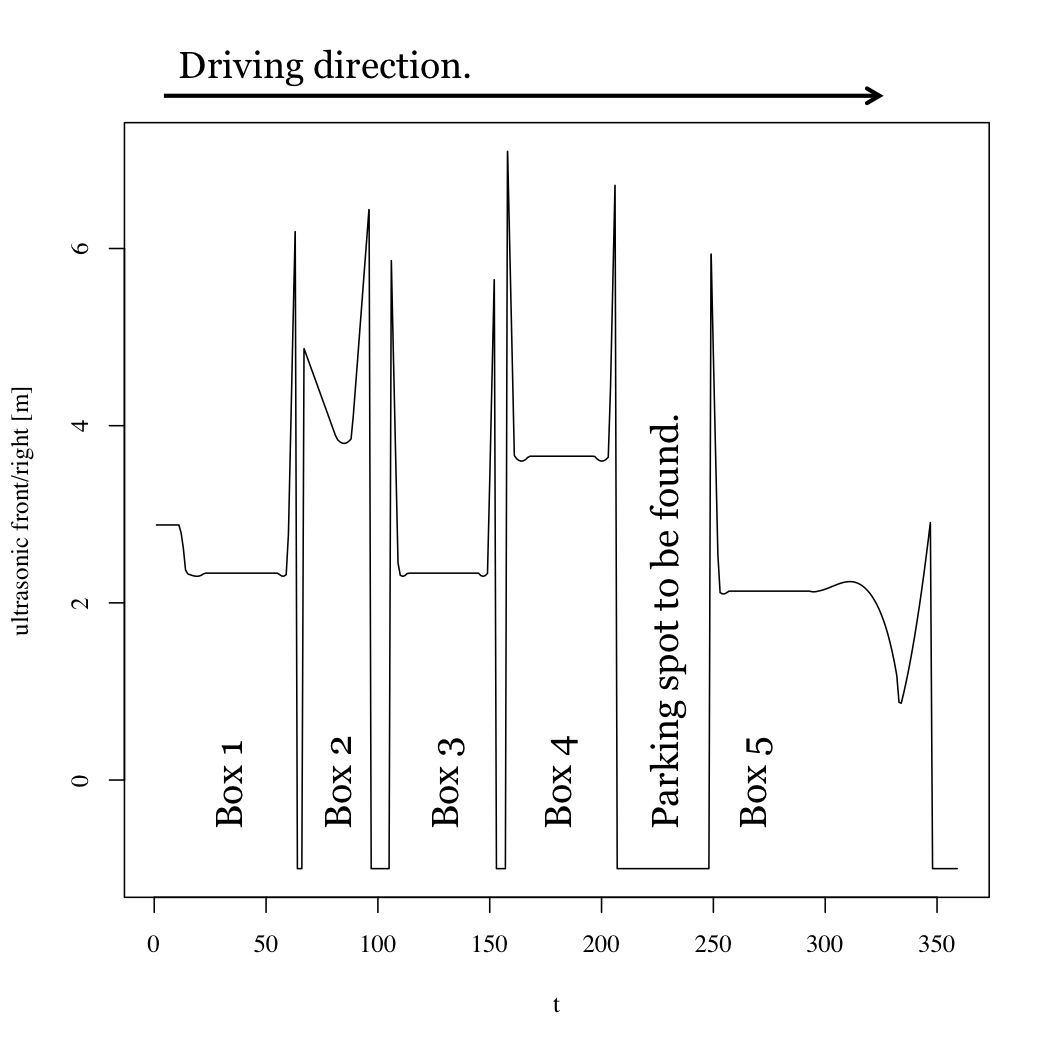}
	        \label{fig:UltrasonicDistancesParkingAnnotations}
	}
	\subfigure[Parking scenario in the simulation environment.]
	{
		\includegraphics[width=.425\linewidth]{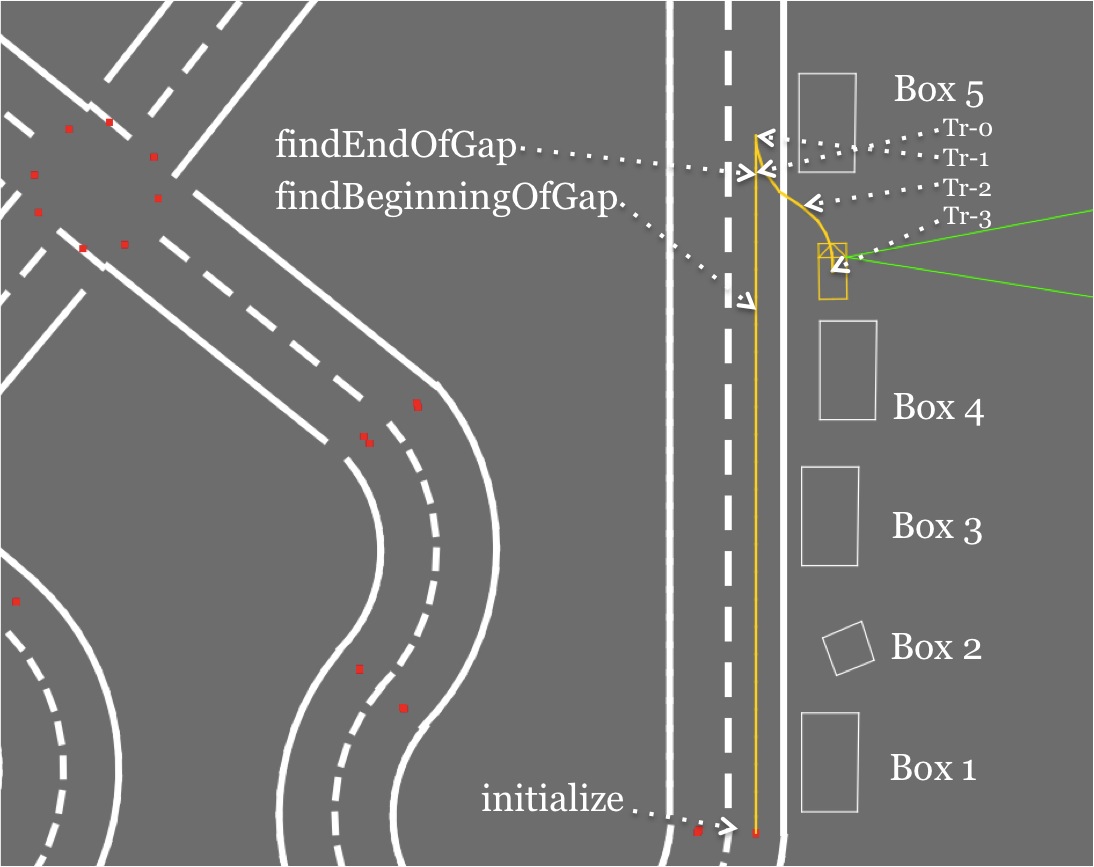}
        	\label{fig:ParkingScenarioParked}
	}
        \caption{Concept for the sideways parking algorithm.}
        \label{fig:SidewaysParking}
  \end{center}  
\end{figure*}

In this section, the design drivers for a state-machine implementing a sideways
parking algorithm is outlined to address \textit{RQ-5}. The focus for this
section is to outline the basic ideas behind sideways parking and its evaluation
in the simulation environment.

\subsection{Design Considerations}

The goal for this algorithm is to handle scenarios as depicted by
Fig.~\ref{fig:ParkingScenarioParked}, where the self-driving miniature vehicle is
placed at the bottom part. After starting the vehicle, it follows the straight
lane while measuring the distances to the obstacles placed on its right hand
side imitating a sideways parking strip.

Once the vehicle has found a parking spot, which is sufficiently wide enough, it
shall stop and move into the parking spot without touching the surrounding
obstacles. Furthermore, the vehicle's heading error in comparison to the straight
road must be less or equal than 5$^\circ$ and the minimum distance to the front and
rear vehicle must be greater or equal than 1cm. Fig.~\ref{fig:ParkingScenarioParked}
also shows that the obstacles on the parking strip can have different dimensions and
can be placed with varying distances to the right lane marking.

The basic idea behind the sideways parking algorithm is shown in
Fig.~\ref{fig:SidewaysParking}. Besides the actual lane-following functionality,
the vehicle's odometer to measure the vehicle's travelled distance over time and
distance data from the ultrasonic sensor, which is mounted at the vehicle's
front/right corner, is used. For the parking scenario shown in
Fig.~\ref{fig:ParkingScenarioParked}, the measured distances over time for
the ultrasonic sensor front/right are depicted in
Fig.~\ref{fig:UltrasonicDistancesParkingAnnotations}. When the sensor does
not measure anything or only obstacles, which are out of the defined viewing
distance, $d_U = -1$ is returned. Therefore, the concept for identifying a parking gap,
which is sufficiently wide enough, is to observe the following event sequence
$d_U > 0 \rightarrow d_U < 0$ followed by $d_U < 0 \rightarrow d_U > 0$
and to measure the driven distance in between. Once the driven distance is
greater or equal than a predefined threshold i.e.~the vehicle's length extended
by an acting margin, the vehicle can terminate the search phase.

After having found the parking gap, there are two possibilities to continue:
(a) the conservative approach is to take this first possible parking gap and to
park the vehicle; (b) another approach is to continue finding a better parking
spot in terms of a more narrow one according to the official rules and
regulations to earn more points in the competition. In the latter case, the
currently found parking spot needs to be saved if no better spot can be found.
Hereby, saving means to record the travelled distance \textit{after} the spot
has been found to be able to return to it later. In any of both cases, a
termination criterion needs to be defined to abort the search for the first or a
better parking gap. This criterion can involve the beginning of a curve since
according to the official rules and regulations document, the parking strip
is located only on the initial part of the track. As an alternative, the totally
travelled distance $D_{total}$ can also be used; however, this $D_{total}$
needs to be determined empirically beforehand.

\begin{figure*}[htb]
  \begin{center}
	\includegraphics[width=.85\linewidth,trim=1cm 1.5cm 1cm .8cm]{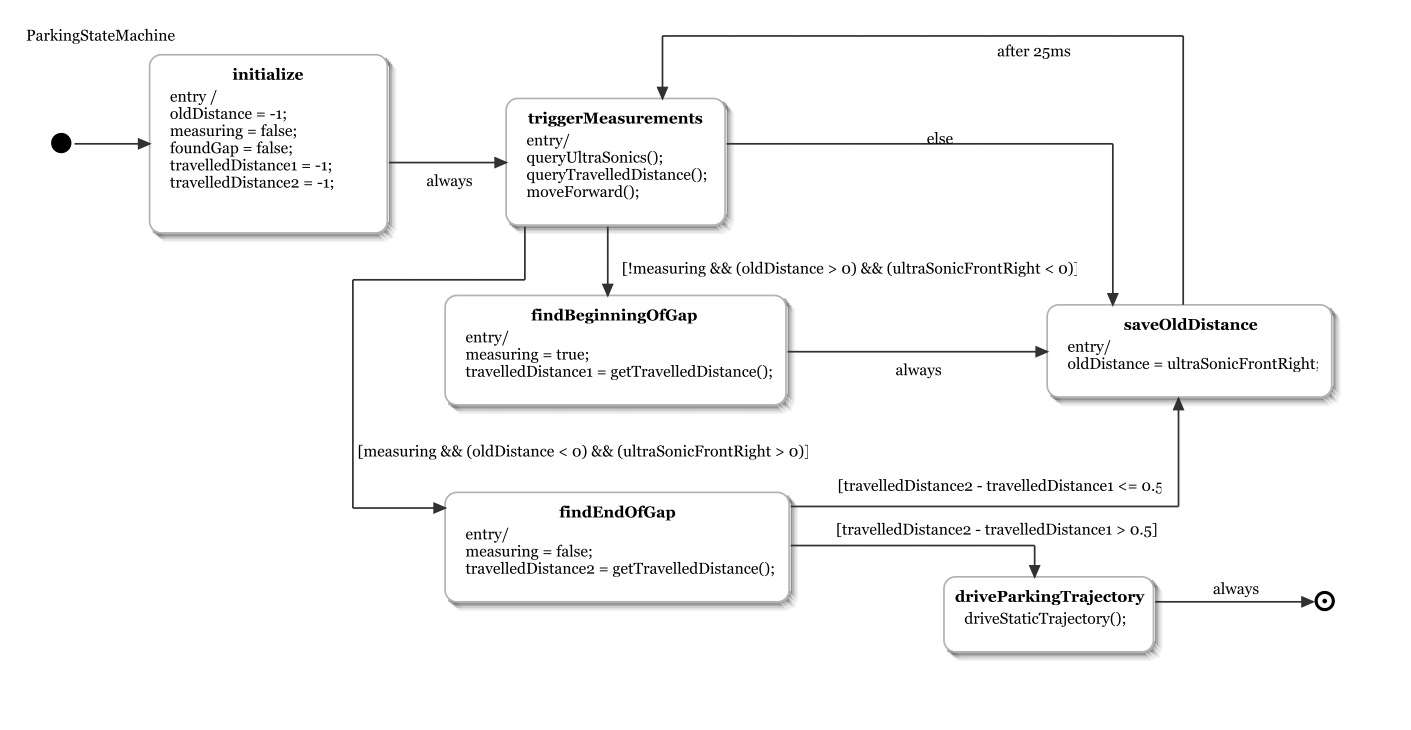}
        \caption{Parking state machine.}
        \label{fig:ParkingStateMachine}
  \end{center}  
\end{figure*}

After a parking spot has been found, the vehicle is stopped to initiate the
parking procedure. This can also be realized in two ways: (a) a predefined
parking trajectory can be ``replayed'', which has been determined empirically
beforehand, or (b) the steering and acceleration parts for the trajectory are
computed online depending on the size of the parking spot. The advantage of
the former is that the movements, which are required to compose a complete
parking trajectory, can be determined independently from the algorithmic
part of finding a proper parking gap. However, the disadvantage is that if the
vehicle is not perfectly aligned after stopping once a sufficiently wide enough
parking gap has been found, the final position and heading of the vehicle might
not be optimal. Therefore, a combination of both where a static trajectory is
adapted according to the current parking situation is recommended.

\subsection{Sideways Parking State-Machine}

Fig.~\ref{fig:ParkingStateMachine} shows the overall state-machine to
realize the aforementioned algorithmic concept with a static parking trajectory.
The main aspect handled by this state-machine is the determination of the size
of an identified parking gap. The outlined state-machine runs with a frequency
of 40Hz to calculate the travelled path while the ultrasonic sensor is sampled
internally every 70ms.

At the beginning, two variables to record the travelled distance at two
consecutive time points are initialized. Afterwards in the state
\texttt{triggerMeasurements}, the sensor as well as the odometer are queried
while the vehicle continuously moves forward. As outlined before, the
fundamental lane-following algorithm can be reused to support this task.

Within this state, it is continuously checked for the event $d_U > 0 \rightarrow d_U < 0$,
which indicates the end of an object. Once this event occurs, the travelled
distance is saved and the boolean measuring flag is set in state
\texttt{findBeginningOfGap} to observe the subsequent event $d_U < 0 \rightarrow d_U > 0$.
Afterwards, the vehicle continues while waiting for the second part of the event
sequence to fire the transition to state \texttt{findEndOfGap}.

This state now ends the measuring phase by inverting the boolean flag.
Additionally, it compares the travelled distance at this time point with the
previous one. If the difference is not large enough to fit into the parking gap,
the state-machine starts over and continues to find another parking spot.
Otherwise, the vehicle is stopped and moved forward for a short distance to
optimize the initial position when starting to drive the static parking
trajectory (part between \texttt{Tr-0} and \texttt{Tr-1}). Afterwards, the
vehicle initiates the parking process with moving backwards while steering
at maximum to the right. Once the vehicle center is half-way in the parking
gap (\texttt{Tr-2}), the vehicle steers at maximum to the left while already
reducing its velocity to come to a full stop at the end (\texttt{Tr-3}).

The concrete values for \texttt{Tr-0}, \texttt{Tr-1}, \texttt{Tr-2} and
\texttt{Tr-3} need to be determined properly. These values can be calculated
analytically with simulative data as well, however, the real values to be used
on the real vehicle are also influenced by the concrete components for the motor
and steering servo and thus, needs to be validated in a real experimental
setting afterwards.

\section{Best Practices and Lesson's Learnt}
\label{sec:BestPractices}

This section summarizes the essential findings from the aforementioned sections.
In this regard, architectural considerations are recapped followed by a description
of the development and evaluation process that was used to develop the self-driving
miniature car ``Meili''.

\subsection{Architectural Considerations}

\emph{Standardized hardware components} allow focusing on integrating the
required components like sensors and actuators. Using COTS components also
reduces the dependency on time-consuming PCB assembly, increases the quality
by professionally assembled boards, and potentially saves costs. The winner of
the 2006 DARPA Urban Challenge, Sebastian Thrun, summarizes this strategy
by saying ``It's all in the algorithms'' (cf.~\cite{Rus06}).

\emph{Standardized hardware abstraction layer} compensates the dependency on
specific COTS components by encapsulating lower layers. In this regard, a
standardized low-level software layer hides implementation details from higher
layers and enables the possibility to change hardware components later.
Furthermore, reusability for software components on the higher layer is
facilitated by providing a standardized software interface. This strategy is also
realized with AUTOSAR \cite{HSFB+04} in real scale automotive platforms.

\emph{Standardized software interfaces} for the software components that are
processing the data on the higher layers enable their reuse in further contexts like
a simulation environment. If the components have cleanly designed interfaces
as exemplified and discussed in \cite{Ray03} to allow the controlled interruption
of communication and execution, a coordinated execution in a virtual test
environment is possible to automatically test the implementation. Furthermore,
cleanly defined software interfaces enable code generation to reduce manual
implementation tasks and to reuse artifacts in a model-based development
process.

\emph{Platform-independent data structures} 

While a standardized hardware abstraction layer reduces the dependency on
specific hardware components, the benefits from such layer needs to be preserved
by using platform-independent data structures. Instead of propagating sensor
properties to the highest layer for example, a generic representation should be
chosen to enable a replacement of the content of such data structures by
synthetically generated data from simulations. Thus, virtual test environments
allow the systematic analysis of an algorithm's behavior and robustness by inject
faulty or noise data for example.

\subsection{Development and Evaluation Process}

The analysis of the team concept presentations in
Sec.~\ref{sec:TechnicalConceptMapping} did not unveil the use of a
specific development process, which was reported to be successful. Therefore,
the approach applied during the development of the 2013 competition car
``Meili'' from Chalmers $|$ University of Gothenburg, Sweden is briefly outlined
in the following.

The applied development process relied mainly on the clear separation between
the actual algorithm conception, hardware and software design, implementation,
evaluation, and the software/hardware integration phase \cite{BR12a}. The
motivation for this separation originated from the hardware design and
purchasing process, which was on the critical project path, and hence, a
time-limiting factor in a sequential development process.

To realize this separation, the simulation environment was used during the
conception, design, development, and evaluation phase of the software for
the self-driving miniature vehicle \cite{BCH+13} with two main purposes:
(a) iteratively develop and interactively validate algorithmic concepts, and
(b) decouple the hardware manufacturing from the software development.
In this regard, the selected development process also addressed some of
today's challenges in automotive software engineering \cite{Bro05}.

The use of the simulation-based development process was embedded in an
agile approach, where the three vehicle functions were divided into smaller work
packages with weekly deliverables and aims. Thus, it was possible to track both,
the progress and algorithmic robustness over time by comparing the behavior in
the simulation from one week to another.

Furthermore, during the integration of the software with the hardware, the
algorithms were adjusted as required by real world challenges. The improved
components were validated in the simulation environment again to preserve
the existing functionality.

\section{Conclusion}
\label{sec:Conclusion}

This article provides concepts, models, and an architecture design for the
software and hardware towards a standardized 1/10 scale vehicle experimental
platform. Therefore, the article analyzed results from a systematic literature
review (SLR), where relevant related work was searched in four digital libraries
and in Google Scholar. This review yields that no study exists so far, how an
experimental platform for miniature vehicles should be designed covering 
general design considerations for software and hardware architecture,
incorporation of a simulative approach, and fundamental algorithmic concepts.

To extend the results of the literature review, the international competition
for self-driving miniature vehicles was systematically reviewed with respect to
results for different disciplines over the last five years. Furthermore, technical
concepts from the 2013 participants were analyzed and mapped to a technical
concept matrix. This mapping unveiled that the most important aspect
for a self-driving vehicle is a robust and reliable lane-following capability as
the basis for further functionalities like overtaking or finding a parking spot.
Furthermore, the competition results showed that having a reliably running
car in terms of robustness of algorithmic approaches is more important than
focusing solely on speed in the competition.

Based on these results and the own experience from participating with a team
in this competition, recommendations for a hardware architecture and a
simulation-supported software architecture are described. Furthermore,
concepts and state-machines for an image processing-based lane-following
algorithm including overtaking capabilities and a sideways parking algorithm
are described that comprise the basic features for the self-driving miniature
cars in the competition.

Nowadays, vehicular functionalities are getting more and more complex because
further information from the vehicle's surroundings is perceived by sensors.
Moreover, these systems are getting interconnected to enable safer and efficient
vehicle systems in terms of vehicle fleets for example. However, pure digital
simulations on the one hand are not considering enough real-world effects and
real-scale experiments on the other hand are too costly in terms of time and
resources, when experiments with theses systems need to be conducted.

In this regard, further use cases for a miniature vehicular experimental
platform as outlined in this article and future work are for example:
Performance analysis of algorithms for self-driving vehicles, validations of
maneuver protocols between an intelligent vehicle and its stationary
surroundings, but also the investigation of advantages and drawbacks of
complex maneuver protocols where several dynamically moving actors in a
traffic situation are involved.

\section*{Acknowledgments}

The author would like to thank the team from Chalmers $|$ University of
Gothenburg who participated with team ``DR.Meili'' and the students and
teaching assistants from the 2013 DIT-168 project course.



\begin{IEEEbiography}[{Berger_20140429_f19}]{Christian Berger} is assistant
professor in the Department of Computer Science and Engineering at Chalmers
$|$ University of Gothenburg, Sweden. He received his
Ph.D. from RWTH Aachen University, Germany in 2010 for his work on challenges
for the software engineering for self-driving cars. He received the
``Borchers'' award in 2011 for his dissertation. He coordinated the
interdisciplinary project for the development of the self-driving car
``Caroline'', which participated in the 2007 DARPA Urban Challenge Final
in the US. His research expertise is on simulative approaches
and model-based software engineering for cyber-physical systems. He published more
than 50 peer-reviewed articles in workshops, conferences, journals, and books.
\end{IEEEbiography}

\vfill

\enlargethispage{-5in}

\end{document}